\documentclass{article}

\usepackage{arxiv}

\usepackage[utf8]{inputenc} 
\usepackage[T1]{fontenc}    
\usepackage{url}            
\usepackage{booktabs}       
\usepackage{amsfonts}       
\usepackage{amsmath,amssymb}
\usepackage[dvipsnames]{xcolor}
\usepackage{nicefrac}       
\usepackage{microtype}      
\usepackage{lipsum}
\usepackage{fancyhdr}       
\usepackage{graphicx}       
\graphicspath{{media/}}     
\usepackage{pifont}
\newcommand{\xmark}{\ding{55}}%

\usepackage{multirow}
\usepackage[position=below]{subfig}
\usepackage{graphicx,subfig}
\usepackage[outline]{contour}
\usepackage{arydshln}

\usepackage[pagebackref=true,breaklinks=true,colorlinks,bookmarks=false]{hyperref}

\title{Jigsaw-ViT: Learning Jigsaw Puzzles in Vision Transformer}

\author{
  Yingyi Chen \\
  ESAT-STADIUS \\
  KU Leuven, Belgium \\
  \texttt{yingyi.chen@esat.kuleuven.be} \\
  \And
  Xi Shen \\
  Tencent AI Lab \\
  Shenzhen, China \\
  \texttt{tisonshen@tencent.com} \\
  \AND 
  Yahui Liu \\
  University of Trento \\
  Trento, Italy \\
  \texttt{yahui.liu@unitn.it} \\
  \And
  Qinghua Tao \\
  ESAT-STADIUS \\
  KU Leuven, Belgium \\
  \texttt{qinghua.tao@esat.kuleuven.be} \\
  \And
  Johan A.K.~Suykens \\
  ESAT-STADIUS \\
  KU Leuven, Belgium \\
  \texttt{johan.suykens@esat.kuleuven.be} \\
}

\begin{document}

\maketitle

\begin{abstract}
    The success of Vision Transformer (ViT) in various computer vision tasks has promoted the ever-increasing prevalence of this convolution-free network.
    The fact that ViT works on image patches makes it potentially relevant to the problem of jigsaw puzzle solving, which is a classical self-supervised task aiming at reordering shuffled sequential image patches back to their original form. Solving jigsaw puzzle has been demonstrated to be helpful for diverse tasks using Convolutional Neural Networks (CNNs), such as feature representation learning, domain generalization and fine-grained classification. 
    \\
    In this paper, we explore solving jigsaw puzzle as a self-supervised auxiliary loss in ViT for image classification, named Jigsaw-ViT. We show two modifications that can make Jigsaw-ViT superior to standard ViT: discarding positional embeddings and masking patches randomly. Yet simple, we find that the proposed Jigsaw-ViT is able to improve on both generalization and robustness over the standard ViT, which is usually rather a trade-off. Numerical experiments verify that adding the jigsaw puzzle branch provides better generalization to ViT on large-scale image classification on ImageNet. Moreover, such auxiliary loss also improves robustness against noisy labels on Animal-10N, Food-101N, and Clothing1M, as well as adversarial examples. 
    Our implementation is available at \url{https://yingyichen-cyy.github.io/Jigsaw-ViT}.
\end{abstract}


\section{Introduction}
\label{sec:introduction}

\begin{figure}[t]
	\centering    
	{\includegraphics[width=0.7\textwidth, height=0.46\textwidth]{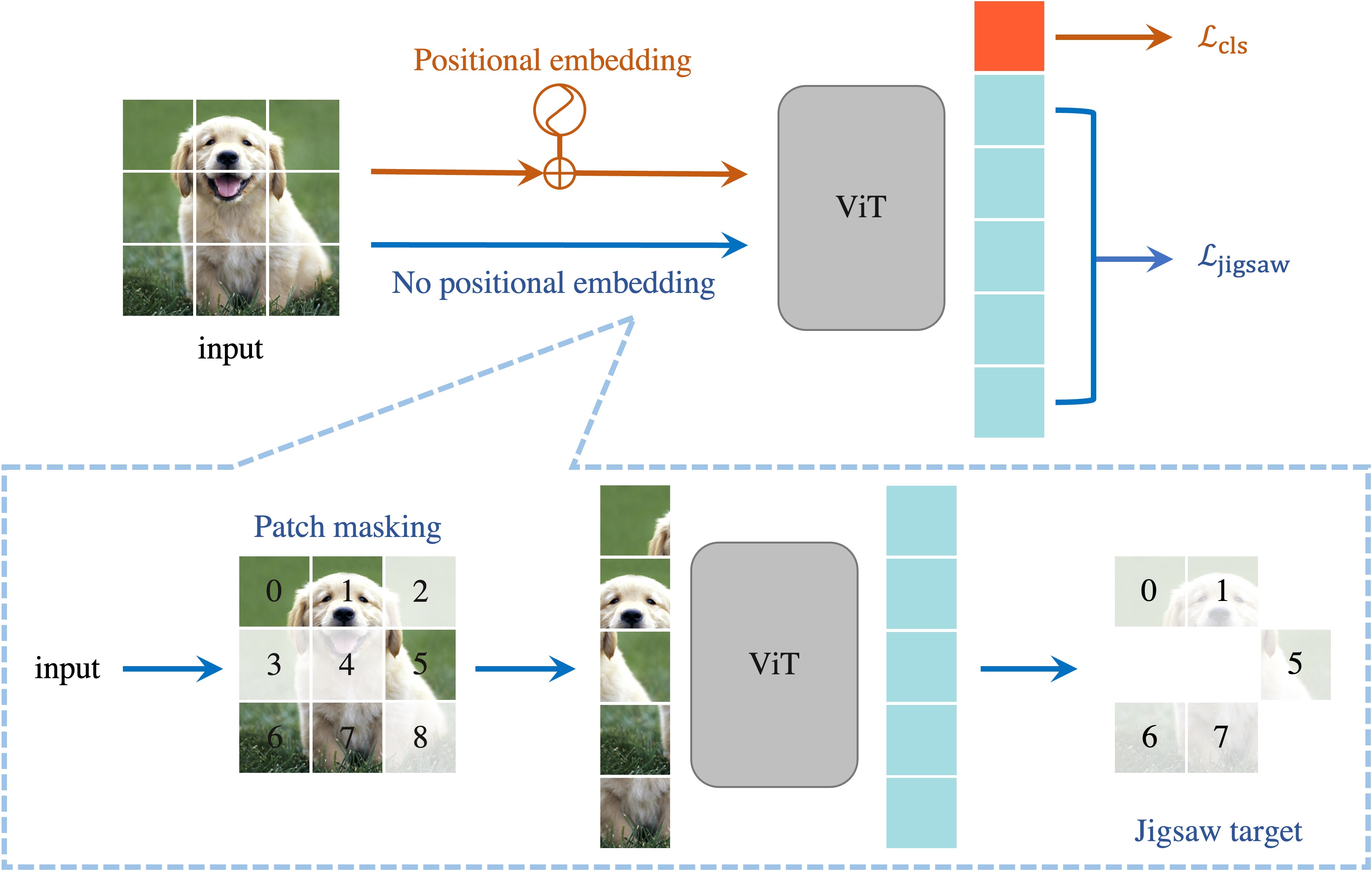}\label{fig:jigsaw}}\hfill
	\caption{\textbf{Overview framework of our Jigsaw-ViT.} 
	(Top) We incorporate jigsaw puzzle solving (in \textbf{\textcolor{NavyBlue}{blue}} flow) into the standard ViT for image classification (in \textbf{\textcolor{BrickRed}{red}}). During the training, we jointly learn the two tasks.
	(Bottom) The details of our jigsaw puzzle flow. We drop several patches, i.e.,~patch masking, and remove positional embeddings before feeding to ViT. For each unmasked patch, the model should predict the class corresponding to the patch position.}
	\label{fig::overview}
\end{figure}
	    
Vision Transformer (ViT)~\cite{dosovitskiy2021an} 
is an architecture inherited from Natural Language Processing~\cite{vaswani2017attention} while applied to image classification with taking raw image patches as inputs. 
Different from classical Convolutional Neural Networks (CNNs), the architectures of ViTs are based on self-attention modules~\cite{vaswani2017attention}, which aim at modeling global interactions of all pixels in  feature maps. 
More precisely, ViTs take sequential image patches as inputs, and the attention mechanism enables  interaction and aggregation directly among patch information. 
Therefore, compared to CNNs where image features are progressively learnt from local to global context via reducing spatial resolution, ViT enjoys obtaining global information from the very beginning. 
Up till now, such convolution-free networks have been achieving great success on various computer vision tasks, including 
{image classification~\cite{touvron2021training,wu2021cvt,chen2021crossvit,li2022contextual,mao2022towards,yao2022wave}, object detection~\cite{liu2021swin,li2022contextual,yao2022dual}, semantic segmentation~\cite{strudel2021segmenter,liu2021swin,yao2022dual}} and image generation~\cite{chen2021pre}, etc.

The fact that ViTs work on image patches makes it potentially relevant to one classical image patch-based learning task, that is, jigsaw puzzle solving. 
Solving jigsaw puzzle aims at reordering shuffled sequential image patches back to their original form. 
In practice, the problem is interesting for cultural heritage and archaeology to search the correct configuration given numerous fragments of an art masterpiece~\cite{paumard2018image}. 
However, in the Computer Vision community, the most interesting aspect could be that it provides \textit{off-the-shelf} annotations for free
considering a given image. 
Despite its simplicity, it has shown effectiveness in diverse Computer Vision tasks based on CNNs such as: self-supervised feature representation learning~\cite{noroozi2016unsupervised}, domain generalization~\cite{carlucci2019domain} and fine-grained classification~\cite{du2020fine}. Motivated by the fact that both jigsaw puzzle solving and ViT share the same basis of learning from image patches, we consider incorporating solving jigsaw puzzle to ViT for image classification tasks.

In this paper, we explore leveraging the jigsaw puzzle solving problem as a self-supervised auxiliary loss of a standard ViT, named Jigsaw-ViT.
Precisely, as shown in Fig.~\ref{fig::overview}, in addition to the standard classification flow in the end-to-end training, we add a jigsaw flow whose goal is to predict the absolute positions of the input patches by solving a classification problem.
Notably, we make two important modifications compared to the naive jigsaw puzzle when feeding input patches to ViTs:
\textit{i)} we get rid of the positional embeddings in the jigsaw flow, by which we prevent the model from cheating from explicit clues in the positional embeddings;
\textit{ii)} we randomly mask some input patches, i.e., \emph{patch masking}, and then aim at predicting only the positions of those unmasked patches, hence making the prediction rely on global context rather than several particular patches.

Despite its simplicity, we find that our Jigsaw-ViT is able to improve on both generalization and robustness over the standard ViT, which is usually rather a trade-off~\cite{zhang2019theoretically}. 
To be specific, in terms of generalization, we observe a steady increase in classification accuracy on ImageNet{-1K}~\cite{deng2009imagenet} that our jigsaw flow brings to the ViTs.
As for robustness, we first show that the proposed jigsaw flow provides consistent improvement against noisy labels on three important real-world benchmarks, i.e.,~Animal-10N \cite{song2019selfie}, Food-101N \cite{lee2018cleannet} and Clothing1M~\cite{xiao2015learning}.
Then, we show that our proposed Jigsaw-ViTs can effectively enhance the robustness of ViTs against adversarial attacks in both black-box and white-box attack settings.

To summarize, our contributions are as follows: \textit{First}, we propose to introduce the jigsaw puzzle solving task into ViT-based models, namely Jigsaw-ViT, with two techniques: removing positional embeddings, and randomly masking patches.
\textit{Second}, empirical results suggest that our jigsaw flow not only improves the generalization ability of ViTs on large-scale image classification, but also the robustness against label noise and adversarial examples.
Our implementation is available at \url{https://yingyichen-cyy.github.io/Jigsaw-ViT}.

\section{Related work}
\label{sec:related_work}
\paragraph{Solving jigsaw puzzle in CNNs}
Solving jigsaw puzzles aims at recovering an original image from its shuffled patches, which is a classical pattern recognition problem dating back to \cite{freeman1964apictorial}.
Rather than setting the jigsaw puzzle solving as the ultimate goal~\cite{gallagher2012jigsaw}, nowaday works treat solving jigsaw puzzles as a pre-text task to other visual recognition tasks~\cite{carlucci2019domain,du2020fine}.
These methods assume that rich feature representations could be learnt in a self-supervised manner, which would be useful for fine-tuning with task-specific data.
For example, \cite{carlucci2019domain} solves classification and jigsaw together to improve semantic understanding for domain generalization tasks, and \cite{du2020fine} combines jigsaw puzzles and the progressive training for fine-grained classification.
These methods use full sequential image patches and are built in the context of CNNs, while here we randomly mask image patches and introduce jigsaw naturally in the context of ViTs.

\paragraph{Vision Transformers}
Transformer proposed in \cite{vaswani2017attention} originally designed for natural language processing has shown promising performance for Computer Vision tasks~\cite{dosovitskiy2021an,wu2021cvt,chen2021crossvit,li2022contextual}.
Vision Transformers (ViTs)~\cite{dosovitskiy2021an} directly inherit from transformer with image patch sequences as inputs, and have achieved superior performance than their counterpart CNNs for {various tasks \cite{he2021transreid,liu2021swin,strudel2021segmenter,yao2022dual}.
The success of ViTs has also encouraged the emergence of a wide variety of ViT variants \cite{touvron2021training,he2021masked,wu2021cvt,chen2021crossvit}.} 
One of the most representative works is DeiT~\cite{touvron2021training}, which introduces a distillation token and a teacher-student strategy specific to transformers, leading to competitive performance on ImageNet~\cite{deng2009imagenet}. 
{Although ViTs are convolution-free, recent works~\cite{li2022contextual,srinivas2021bottleneck,wang2021pyramid} also build stronger ViT variants by resembling some merits from CNNs.}
Notably, our work is different from \cite{he2021transreid} since \cite{he2021transreid} does not include jigsaw puzzle solving in the optimization goal, where they only shuffle patches in the triplet loss branch.

\paragraph{Learning with noisy labels}
The task aims at learning models achieving good clean test accuracy while being trained on noisy annotated data.
Mainstream solutions include: 
label correction which corrects possibly wrong labels with more consistent substitutes~\cite{tanaka2018joint,zhang2021learning}, 
semi-supervised learning which trains networks in a semi-supervised manner with only the clean labels used \cite{li2020dividemix}, and sample reweighting which assigns more weights to samples possibly clean~\cite{han2018co}.
In particular, for sample reweighting, Co-teaching \cite{han2018co} is a classical method that cross-updates its two base networks on the small-loss samples selected by its peer.
Based on this, Nested Co-teaching \cite{chen2022compressing} improves performance by including compression regularization during the training.

\begin{figure*}[t]
		\centering
		\begin{minipage}{0.1\textwidth}
		\centering
		\footnotesize{Input}
		\end{minipage}\hfill
		\begin{minipage}{0.9\textwidth}
		\centering
		\includegraphics[width=0.1\columnwidth,height=0.1\columnwidth]{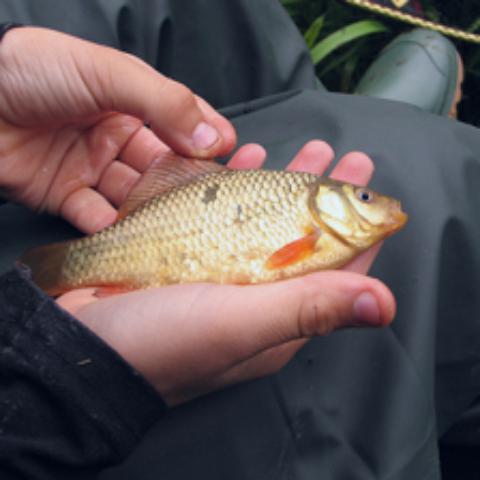}
		\includegraphics[width=0.1\columnwidth,height=0.1\columnwidth]{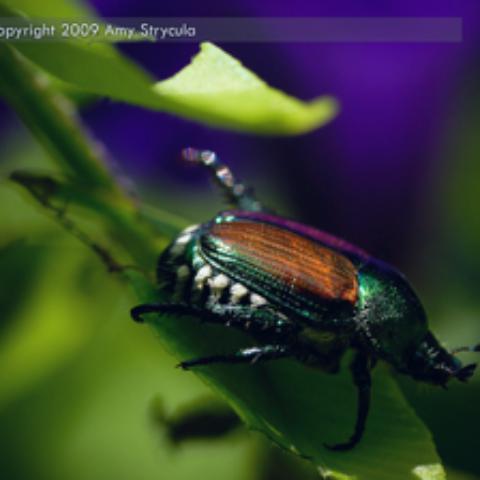}
		\includegraphics[width=0.1\columnwidth,height=0.1\columnwidth]{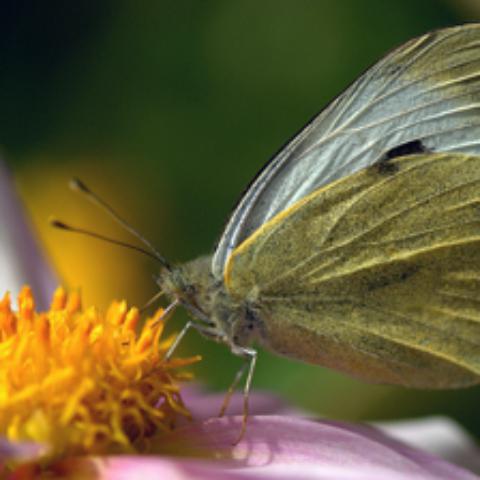}
		\includegraphics[width=0.1\columnwidth,height=0.1\columnwidth]{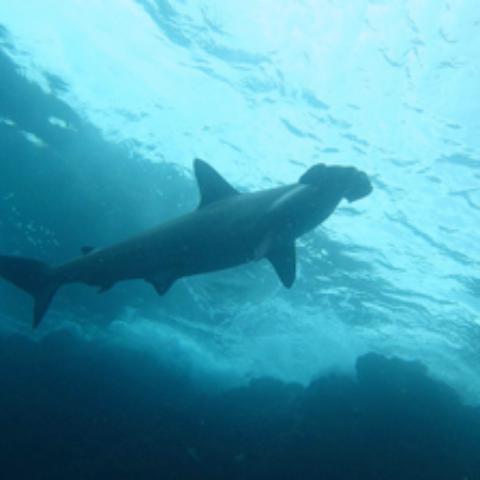}
		\includegraphics[width=0.1\columnwidth,height=0.1\columnwidth]{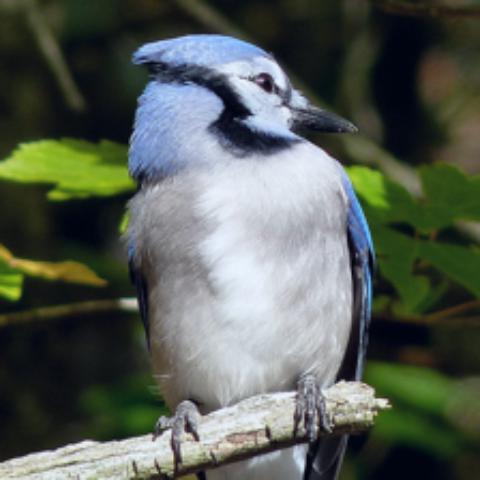}
		\includegraphics[width=0.1\columnwidth,height=0.1\columnwidth]{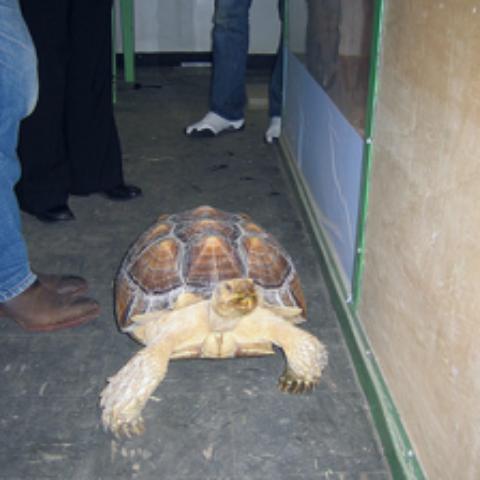}
		\includegraphics[width=0.1\columnwidth,height=0.1\columnwidth]{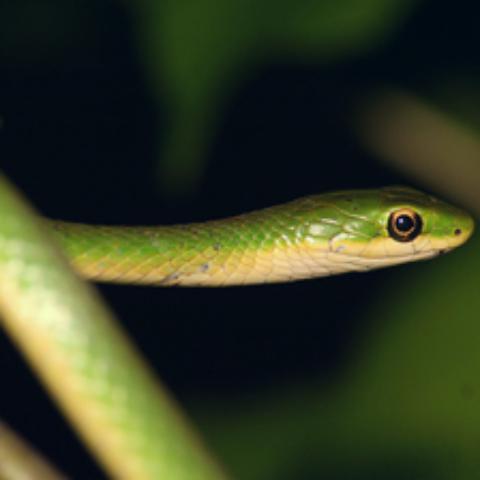}
		\includegraphics[width=0.1\columnwidth,height=0.1\columnwidth]{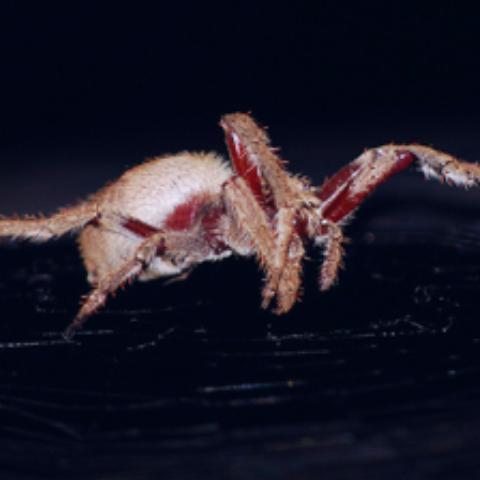}
		\includegraphics[width=0.1\columnwidth,height=0.1\columnwidth]{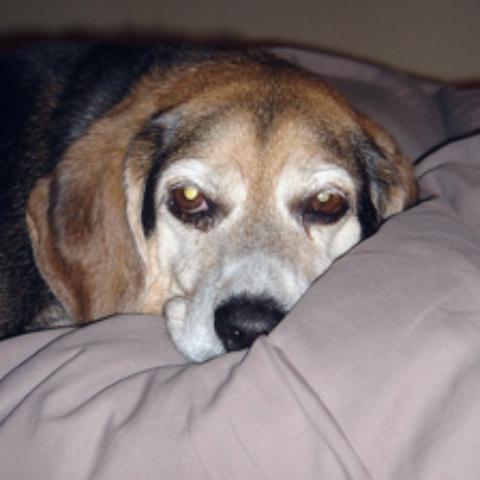}
		\end{minipage}\hfill
		\begin{minipage}{0.1\textwidth}
		\centering
		\footnotesize{DeiT-Small/16}
		\end{minipage}\hfill
		\begin{minipage}{0.9\textwidth}
		\centering
		\includegraphics[width=0.1\columnwidth,height=0.1\columnwidth]{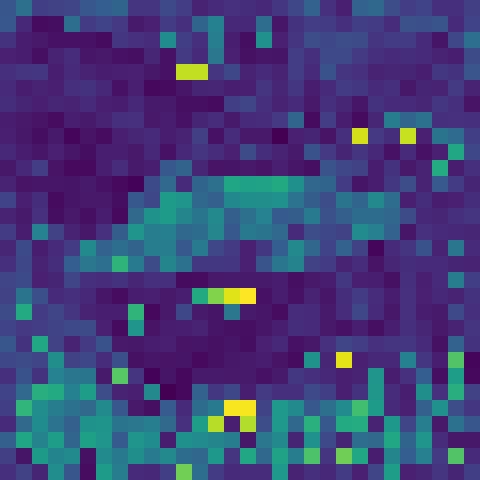}
		\includegraphics[width=0.1\columnwidth,height=0.1\columnwidth]{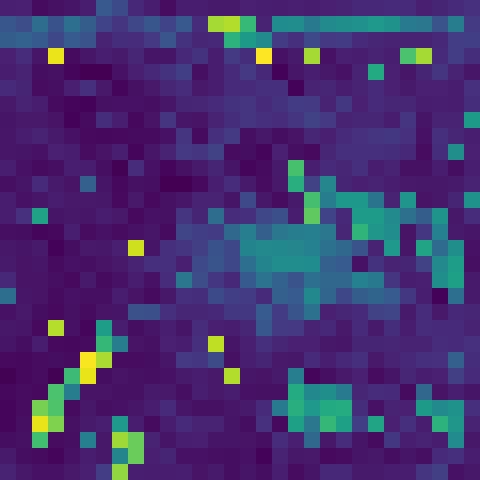}
		\includegraphics[width=0.1\columnwidth,height=0.1\columnwidth]{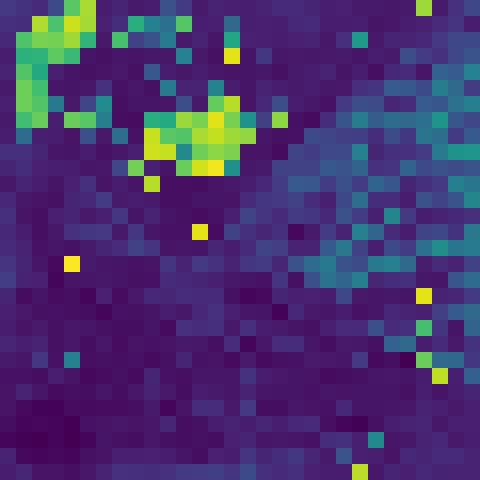}
		\includegraphics[width=0.1\columnwidth,height=0.1\columnwidth]{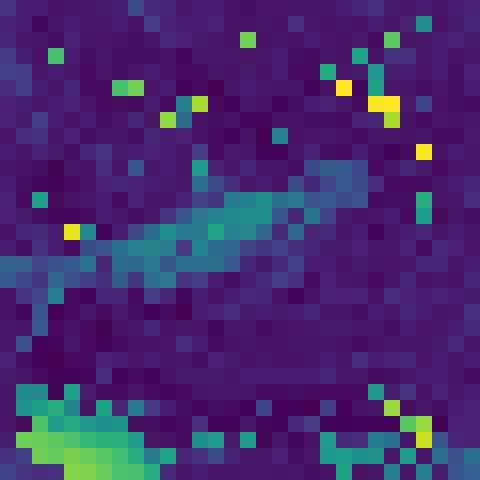}
		\includegraphics[width=0.1\columnwidth,height=0.1\columnwidth]{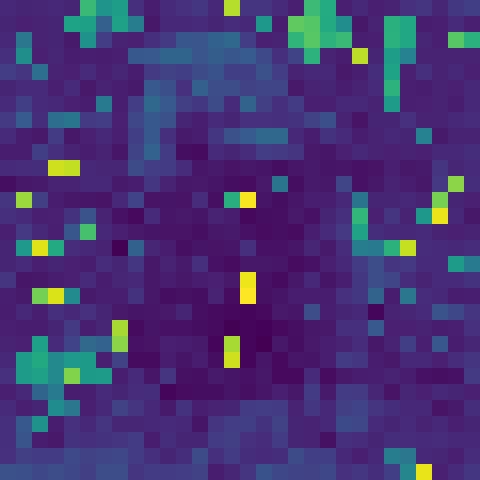}
		\includegraphics[width=0.1\columnwidth,height=0.1\columnwidth]{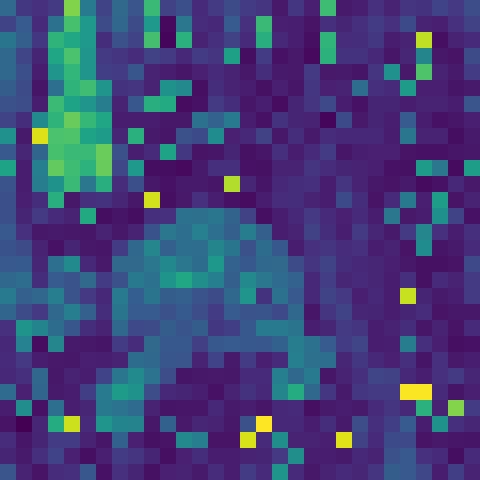}
		\includegraphics[width=0.1\columnwidth,height=0.1\columnwidth]{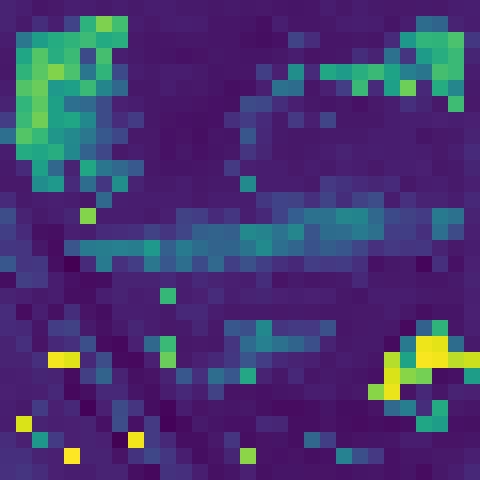}
		\includegraphics[width=0.1\columnwidth,height=0.1\columnwidth]{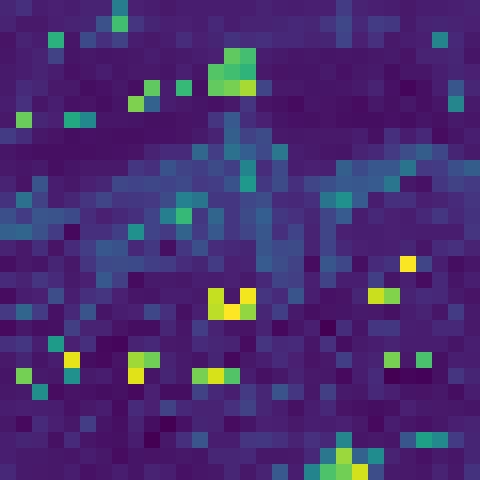}
		\includegraphics[width=0.1\columnwidth,height=0.1\columnwidth]{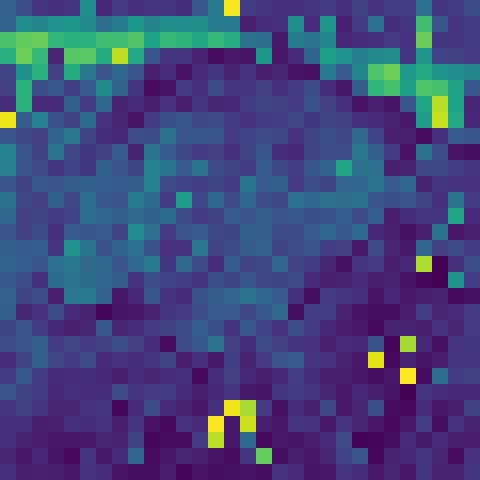}
		\end{minipage}\hfill
		\begin{minipage}{0.1\textwidth}
		\centering
		\footnotesize{Jigsaw-ViT}
		\end{minipage}\hfill
		\begin{minipage}{0.9\textwidth}
		\centering
		\includegraphics[width=0.1\columnwidth,height=0.1\columnwidth]{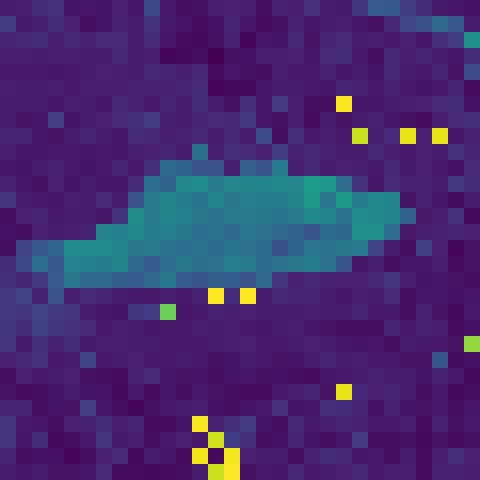}
		\includegraphics[width=0.1\columnwidth,height=0.1\columnwidth]{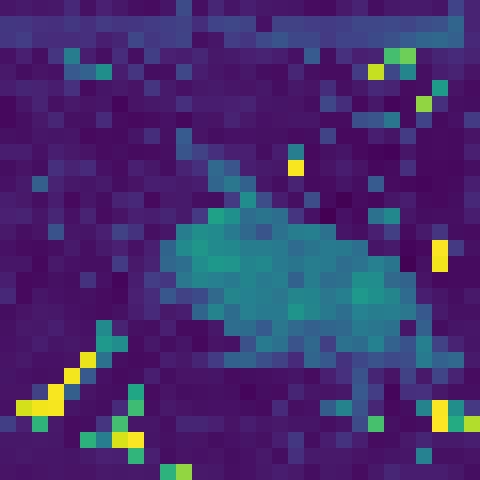}
		\includegraphics[width=0.1\columnwidth,height=0.1\columnwidth]{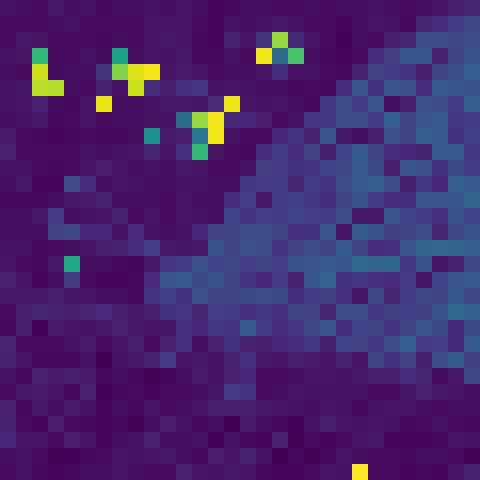}
		\includegraphics[width=0.1\columnwidth,height=0.1\columnwidth]{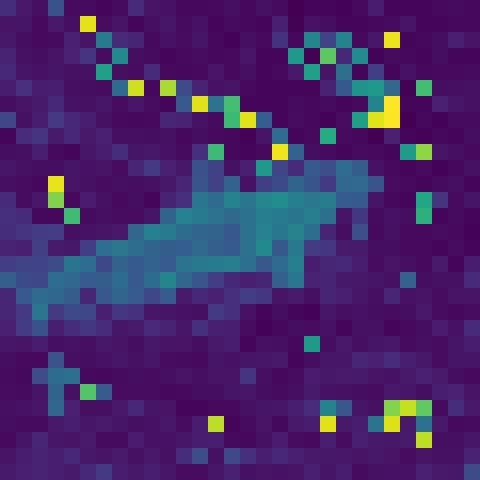}
		\includegraphics[width=0.1\columnwidth,height=0.1\columnwidth]{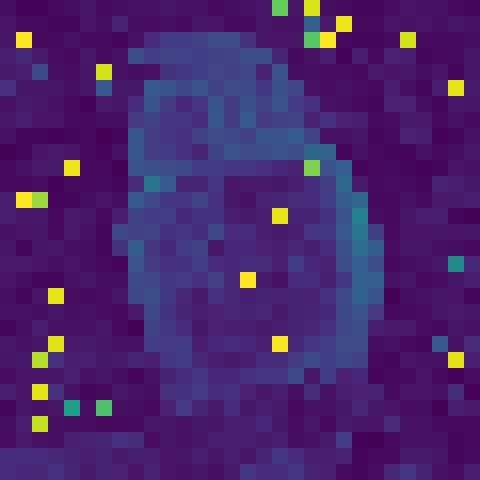}
		\includegraphics[width=0.1\columnwidth,height=0.1\columnwidth]{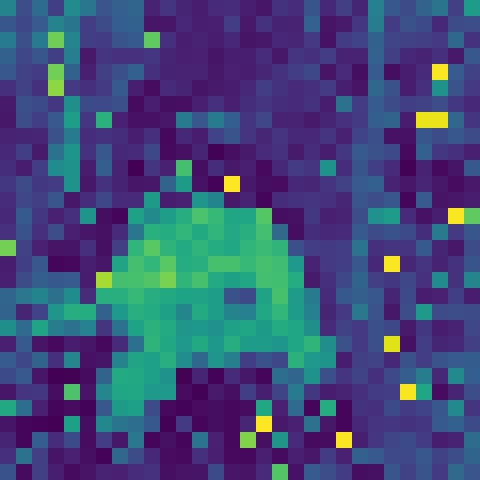}
		\includegraphics[width=0.1\columnwidth,height=0.1\columnwidth]{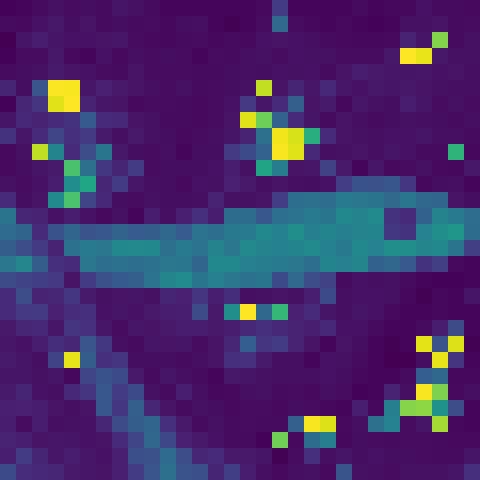}
		\includegraphics[width=0.1\columnwidth,height=0.1\columnwidth]{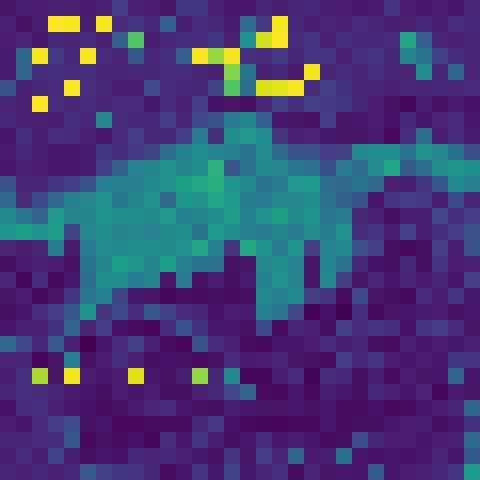}
		\includegraphics[width=0.1\columnwidth,height=0.1\columnwidth]{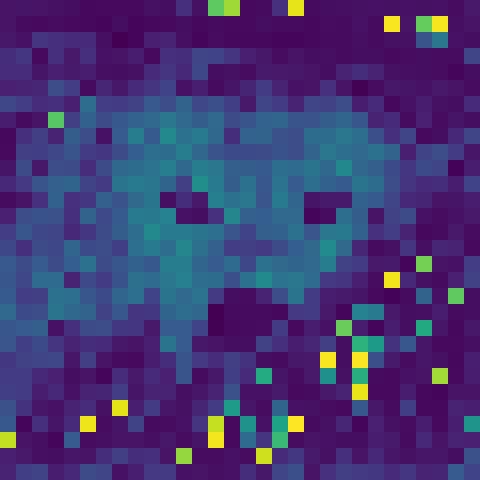}
		\end{minipage}
		\caption{{\bf Attention map associated to the class token of the last layer.} We show the attention map for DeiT-Small/16~\cite{touvron2021training} and Jigsaw-ViT trained on ImageNet{-1K} \cite{deng2009imagenet}. 
		Jigsaw-ViT learns clearer salient-object attentions over the listed instances.
		More examples can be found in the supplementary material.}
		\label{fig::attention}
	\end{figure*}
	
\paragraph{Adversarial examples}
Deep neural networks are fragile to adversarial examples \cite{dalvi2004adversarial} where human-imperceptible perturbations on clean images can cheat the network to give wrong predictions. 
These adversarial examples can be generated in the white-box settings where attacker has full access to information inside
the target model.
Mainstream attacks include Fast Gradient Sign Method (FGSM)~\cite{goodfellow2014explaining}, projected gradient descent (PGD)~\cite{madry2018towards}, the ensemble auto-attack (AA)~\cite{croce2020reliable}, etc.
In real-world scenarios, adversarial attacks are commonly done in black-box settings since the much information, e.g., gradients, of the target models are hard to obtain.
Existing black-box attacks are mostly conducted either in query-based~\cite{yang2020learning} or transfer-based ways~\cite{papernot2017practical}.
The former relies on querying the outputs of the target models, while the latter uses surrogate models to generate adversarial examples.
Recent work~\cite{mahmood2021robustness} also studies the adversarial robustness of ViTs where an ensemble of ViTs and CNNs can achieve good robustness.

\section{Method}
\label{sec:method}
In this section, we present details of the proposed Jigsaw-ViT. 
A brief introduction to ViT~\cite{dosovitskiy2021an} is firstly given in Section~\ref{subsec::ViT}. 
Then, we present our Jigsaw-ViT in Section~\ref{subsec::jigsaw}. 

\subsection{Vision Transformer}
\label{subsec::ViT}

Given an image $\mathbf{I} \in \mathbb{R}^{H \times W \times C}$, where $H$ and $W$ are spatial dimensions and $C$ denotes the number of channels, we first divide the image into a sequence of non-overlapped 2D patches $\mathbf{I}_p \in \mathbb{R}^{L \times (P \times P \times C)}$ where the patch resolution is $P \times P$ and $L = HW/P^2$ is the sequence length (the number of patches). 
In ViT~\cite{dosovitskiy2021an}, these patches are linearly projected to $D$-dimensional features used as the patch embeddings $[\mathbf{z}_0^1,\text{ }\mathbf{z}_0^2\text{ }...\text{ }\mathbf{z}_0^L]^T\in \mathbb{R}^{L \times D}$. 
A learnable class token denoted by $CLS$, \textit{i.e.},~$\mathbf{z}_0^{\text{cls}} \in \mathbb{R}^{D}$ is prepended to the sequence of the patch embeddings $\mathbf{z}_0$ leading to $\mathbf{z}_0 =  [\mathbf{z}_0^{\text{cls}},\text{ } \mathbf{z}_0^1,\text{ }\mathbf{z}_0^2\text{ }...\text{ }\mathbf{z}_0^L]^T \in \mathbb{R}^{(L+1) \times D}$. 
Usually, positional embeddings $\mathbf{p} = [\mathbf{p}^{\text{cls}},\text{ }\mathbf{p}^1,\text{ }\mathbf{p}^2\text{ }...\text{ }\mathbf{p}^L]^T \in \mathbb{R}^{(L+1) \times D}$ are added to the sequential patch embeddings, thus $\mathbf{v}_0 = \mathbf{z}_0 + \mathbf{p} $ serves as the input to the Transformer encoder. 

The Transformer encoder~\cite{vaswani2017attention} consists of alternating layers of layer normalization, multi-head self-attention and multi-layer perceptron blocks, denoted as $LN(\cdot)$, $MSA(\cdot)$ and $MLP(\cdot)$, respectively. 
For an encoder with $N$ layers, the final class prediction $\mathbf{y}_{\text{pred}}$ is the final embedding associated to the class token $\mathbf{v}_{N}^{\text{cls}}$, such that

\begin{equation}
    \begin{aligned}
    \mathbf{v}'_i & = MSA(LN(\mathbf{v}_{i-1})) + \mathbf{v}_{i-1}, &i \in [1, 2, ..., N] \\
    \mathbf{v}_i & = MLP(LN(\mathbf{v}'_i)) + \mathbf{v}'_i, &i \in [1, 2, ..., N] \\
    \mathbf{y}_{\text{pred}}& = MLP(LN(\mathbf{v}^{\text{cls}}_{N})).
\end{aligned}
\end{equation}

\subsection{Jigsaw-ViT}
\label{subsec::jigsaw}
    
Solving jigsaw puzzles aims at reordering shuffled sequential patches back to their original format. 
Previous CNN-based methods have proved that various computer vision tasks are benefited from learning jigsaw puzzles~\cite{noroozi2016unsupervised,carlucci2019domain,du2020fine}. 
In this section, we show how to incorporate a jigsaw puzzle flow into the regular end-to-end training of standard ViTs. 
    
The overview of our approach is illustrated in Fig.~\ref{fig::overview} where image classification problem using ViT is the focus in this paper.
More precisely, our goal is to train a ViT model that jointly considers solving the standard classification and jigsaw puzzles in its optimization objective.
Accordingly, the total loss $\mathcal{L}_{\text{total}}$ simultaneously involves two cross-entropy losses ($CE$s), i.e.,~the class prediction loss on the class token $\mathcal{L}_{\text{cls}}$ and the position prediction loss on the patch tokens $\mathcal{L}_{\text{jigsaw}}$:
\begin{align} 
\label{eq::obj_func}
    \mathcal{L}_{\text{total}} = \underbrace{CE(\mathbf{y}_{\text{pred}}, \mathbf y)}_{\mathcal{L}_{\text{cls}}} + \eta \underbrace{CE(\tilde{\mathbf{y}}_{\text{pred}}, \tilde{\mathbf y})}_{\mathcal{L}_{\text{jigsaw}}}
\end{align}
where $\tilde{\mathbf{y}}_{\text{pred}}$ and $\tilde{\mathbf y}$ denote the position prediction and the corresponding real position, respectively, and $\eta$ is a hyper-parameter balancing the two losses.

As detailed in the bottom part of Fig.~\ref{fig::overview}, our injected jigsaw puzzle flow is different from naive jigsaw puzzle implementation in twofold: 
\textit{i)} we get rid of positional embeddings in the jigsaw puzzle flow; 
\textit{ii)} we randomly mask $\lfloor\gamma L\rfloor$ patches in the input where $\gamma\in[0,1)$ is a hyper-parameter denoting the mask ratio. 
The former prevents models from being cheated from explicit clues in the positional embeddings, and the latter encourages the prediction to rely on global context rather than several particular patches. 
Their helpfulness for the main classification task is shown in Section~\ref{sec::experiments}.
Since the proposed jigsaw puzzle solving is a self-supervised task, it can be easily plugged into existing ViTs without modifying the original architecture.

\paragraph{Architecture} 
We employ DeiT-Small/16~\cite{touvron2021training} without distillation token as the backbone in our experiments if not specified, which is the same architecture as ViT-Small/16 in~\cite{dosovitskiy2021an}. DeiT-Small/16 has the embedding dimension of $D=384$ with 6 heads, $N$ = 12 layers and the total number of parameters is approximately 22M. 
The training image resolution is $224\times 224$ leading to 14 $\times$ 14 patches as inputs. 
The prediction head of the image classification flow is a single fully connected layer mapping from the encoder embedding dimension to the number of classes.
For our jigsaw flow, we adopt a 3-layer $MLP$ head after the encoder where the dimensions of the first two layers are equal to the encoder embedding dimension, and the output dimension of the last layer equals to the total number of image patches ($L$ = 14 $\times$ 14).

\section{Experiments}
\label{sec::experiments}

In this section, we demonstrate the effectiveness of our approach on three tasks: generalization on large-scale classification on ImageNet~\cite{deng2009imagenet}, and robustness to noisy labels and adversarial examples. 
In Section~\ref{subsec:imagenet}, we evaluate our approach on the classification task with ImageNet dataset. 
In Section~\ref{subsec::labelnoise}, we extensively validate our approach on three real-world noisy label datasets: Animal-10N~\cite{song2019selfie}, Food-101N~\cite{lee2018cleannet} and Clothing1M~\cite{xiao2015learning}. 
In Section~\ref{subsec::adv}, we show our jigsaw flow improves ViTs' robustness against adversarial examples.
Ablation study is given in Section~\ref{sec::ablation}. 
{Further discussions on setting jigsaw puzzle solving as a pretext task and exploring the benefits of Jigsaw-ViT to downstream tasks are additionally provided in Section~\ref{sec::further}.} 
More experimental details including ablations can be found in the supplementary material.

\subsection{Generalization on large-scale image classification}
\label{subsec:imagenet}

\captionsetup[table]{farskip=2pt,captionskip=1pt,aboveskip=4pt}
    \begin{table}[t]
        \centering
        \renewcommand{\arraystretch}{1.1}
        \caption{{\textbf{Image classification on ImageNet-1K~\cite{deng2009imagenet} validation set {and ImageNet V2~\cite{recht2019imagenet}.}} 
        We compare to DeiT~\cite{touvron2021training} with different capacities and report top-1 accuracy ($\%$). We also show how much each Jigsaw-ViT model is above the baseline with \textcolor{ForestGreen}{$\uparrow$}.}}
        \scalebox{0.8}{
            \begin{tabular}{c|c||cc|cl}
            \multirow{2}{*}{{Backbone}} 
            & \multirow{2}{*}{{\#params}}
            & \multicolumn{2}{c|}{{ImageNet{-1K}}} & \multicolumn{2}{c}{{ImageNet V2}} 
            \\
            & & {Baseline} & {Jigsaw-ViT} & {Baseline} & {Jigsaw-ViT}         
            \\ \hline
            {DeiT-Tiny/16} 
            & {5M}
            & {72.2} 
            & {\bf{74.1} \textcolor{ForestGreen}{$\uparrow 1.9$}}   
            & {60.2}
            & {\bf{61.4} \textcolor{ForestGreen}{$\uparrow 1.2$}}       \\
            {DeiT-Small/16} 
            & {22M}
            & {79.8} 
            & {\bf{80.5} \textcolor{ForestGreen}{$\uparrow 0.7$}}  
            & {68.5} 
            & {\bf{69.3} \textcolor{ForestGreen}{$\uparrow 0.8$}} 
            \\
            {DeiT-Base/16} 
            & {86M}
            & {81.8} 
            & {\bf{82.1} \textcolor{ForestGreen}{$\uparrow 0.3$}}
            & {\bf{71.0}} 
            & {\bf{71.0} }  
            \\ \bottomrule[0.75pt]
            \end{tabular}}
        \label{tab::imagenet}
\end{table}

ImageNet~\cite{deng2009imagenet} is a standard dataset for large-scale image classification, and we use ILSVRC-2012 ImageNet{-1K} dataset containing $1{\small,}000$ classes and approximately 1.3M images for the evaluation of our Jigsaw-ViTs in improving  standard ViTs on large-scale image classification task.

\paragraph{Training details}
We train both DeiT~\cite{touvron2021training} and our Jigsaw-ViT from scratch following the same training protocols in \cite{touvron2021training}, where AdamW is taken as optimizer with the base learning rate of 5e-4, a weight decay of 0.05, the overall batch size as 1,024, and 300 training epochs. 
If not specified, for all experiments, we follow the data augmentation strategies in \cite{touvron2021training}, e.g., Rand-Augment, MixUp and CutMix. 
We set the balancing hyper-parameter in~\eqref{eq::obj_func} $\eta=0.1$ and mask ratio $\gamma=0.5$ here.

\paragraph{Results} 
{Table \ref{tab::imagenet} shows the performances of Jigsaw-ViTs with different backbones on ImageNet{-1K} validation set and ImageNet V2~\cite{recht2019imagenet} which is a distinct test set suitable for measuring the overfitting level of models.
The backbones include DeiT~\cite{touvron2021training} with different capacities from tiny to base.}
DeiT-Tiny/16 is similar to DeiT-Small/16 but with fewer parameters: embedding dimension of 192 with 3 heads, 12 layers and the total number of parameters is approximately 5M. 
{DeiT-Base/16 has an embedding dimension of 768, 12 heads, 12 layers and an approximate total number of parameters of 86M.} 
{For all architectures on both datasets, adding jigsaw branch consistently improves upon the baselines 
by training from scratch or fine-tuning}, verifying the effectiveness of our Jigsaw-ViT in attaining better generalization performances on large-scale image classification. 

To further investigate the impact of our injected jigsaw flow in ViTs, we visualize the self-attention maps of the baseline ViT and Jigsaw-ViT trained on ImageNet{-1K} with DeiT-Small/16 in Fig.~\ref{fig::attention} following the visualization protocols in~\cite{caron2021emerging}, which concatenates features of different heads associated to the class token.
As in Fig.~\ref{fig::attention}, Jigsaw-ViT is able to learn more distinctive salient-object attentions than DeiT-Small/16.
The reason of this difference can be that solving jigsaw puzzle in ViT requires to understand the whole image so as to predict the correct spatial relationship between different image patches with a randomly shuffled order.
Note that there are noisy hightlights on the background for both DeiT-Small/16 and Jigsaw-ViT.
This is consistent with the statement in~\cite{caron2021emerging} that supervised ViTs attend less well to objects in both qualitatively and quantitatively than pure self-supervised ViTs.
However, even though there are noises in attention maps, the proposed Jigsaw-ViT still manages to obtain better attention maps than its counterpart DeiT-Small/16.
We refer to supplementary material for more visual results.

\subsection{Robustness to label noise}
\label{subsec::labelnoise}

\captionsetup[table]{farskip=2pt,captionskip=1pt,aboveskip=4pt}
\begin{table*}[t]
    \centering
    \renewcommand{\arraystretch}{1.1}
    \caption{\textbf{Image classification on datasets with low noise rate.} We compare to state-of-the-art approaches and report test top-1 accuracy ($\%$) on Animal-10N \cite{song2019selfie} (noise ratio $\sim8\%$, in Table (a)) Food-101N~\cite{lee2018cleannet} (noise ratio $\sim20 \%$, in Table (b)). We also show how much Jigsaw-ViT model is above DeiT-Small/16~\cite{touvron2021training} with \textcolor{ForestGreen}{$\uparrow$}. }
        \label{tab::lownoiserate}
        \begin{minipage}{\textwidth}
            \centering
            \scalebox{0.8}
            {\subfloat[Animal-10N \cite{song2019selfie}]{
                \begin{tabular}{c|cccccc||cc}
                \multirow{2}{*}{Method} & \multirow{2}{*}{\begin{tabular}[c]{@{}c@{}}CE\\ {\cite{zhang2021learning}}\end{tabular}} & \multirow{2}{*}{\begin{tabular}[c]{@{}c@{}}Dropout\\ {\cite{chen2021boosting,srivastava2014dropout}}\end{tabular}} & \multirow{2}{*}{\begin{tabular}[c]{@{}c@{}}SELFIE\\ {\cite{song2019selfie}}\end{tabular}} & \multirow{2}{*}{\begin{tabular}[c]{@{}c@{}}PLC\\ {\cite{zhang2021learning}}\end{tabular}} & \multirow{2}{*}{\begin{tabular}[c]{@{}c@{}}NCT\\ {\cite{chen2021boosting, chen2022compressing}}\end{tabular}} & \multirow{2}{*}{\begin{tabular}[c]{@{}c@{}}S3\\ {\cite{feng2021s3}}\end{tabular}} & \multicolumn{2}{c}{\bf Ours} \\
                & & & & & & 
                & DeiT-Small/16~\cite{touvron2021training} & Jigsaw-ViT 
                \\ \hline
                Acc. (\%)               
                & 79.4 & 81.3 & 81.8 & 83.4 & 84.1 & 88.5 
                & 87.2 & {\bf 89.0} \textcolor{ForestGreen}{$\uparrow 1.8$}
                \\ \cdashline{1-9}
                Backbone                
                & \multicolumn{6}{c||}{VGG-19bn~\cite{simonyan2014very}, \#params: 143.7M, FLOPS: 19.7G} & \multicolumn{2}{c}{DeiT-Small/16, \#params: 22M, FLOPS: 4.6G} 
                \\ \bottomrule[0.75pt]
                \end{tabular}\label{tab::animal10}}}
        \end{minipage}\hfill
        \begin{minipage}{\textwidth}
            \centering
            \scalebox{0.8}
            {\subfloat[Food-101N \cite{lee2018cleannet}]{
                \begin{tabular}{c|ccccccc||cc}
                \multirow{2}{*}{Method} & \multirow{2}{*}{\begin{tabular}[c]{@{}c@{}}CE\\ {\cite{zhang2021learning}}\end{tabular}} & \multirow{2}{*}{\begin{tabular}[c]{@{}c@{}}CleanNet\\ {\cite{lee2018cleannet}}\end{tabular}} & \multirow{2}{*}{\begin{tabular}[c]{@{}c@{}}MWNet\\ {\cite{shu2019meta,sun2022learning}}\end{tabular}} & \multirow{2}{*}{\begin{tabular}[c]{@{}c@{}}SMP\\
                {\cite{han2019deep}}\end{tabular}} & \multirow{2}{*}{\begin{tabular}[c]{@{}c@{}}NRank\\
                {\cite{sharma2020noiserank}}\end{tabular}} & \multirow{2}{*}{\begin{tabular}[c]{@{}c@{}}PLC\\ {\cite{zhang2021learning}}\end{tabular}} & \multirow{2}{*}{\begin{tabular}[c]{@{}c@{}}WarPI\\ {\cite{sun2022learning}}\end{tabular}} & 
                \multicolumn{2}{c}{\bf Ours}\\
                &  &  &  &  &  &  &  & DeiT-Small/16~\cite{touvron2021training} & Jigsaw-ViT
                \\ \hline
                Acc. (\%) & 81.7 & 83.5 & 84.7 & 85.1 & 85.2 & 85.3 & 85.9 
                & 84.2 & {{\bf 86.7} \textcolor{ForestGreen}{$\uparrow 2.5$}}
                \\ \cdashline{1-10}
                Backbone                
                & \multicolumn{7}{c||}{ResNet-50~\cite{he2016deep}, \#params:  25.6M, FLOPS: 4G}
                & \multicolumn{2}{c}{DeiT-Small/16, \#params: 22M, FLOPS: 4.6G} 
                \\ \bottomrule[0.75pt]
                \end{tabular}\label{tab::food101}}}
        \end{minipage}\hfill
    \end{table*}

\captionsetup[table]{farskip=2pt,captionskip=1pt,aboveskip=4pt}
    \begin{table*}[t]
    \renewcommand{\arraystretch}{1.1}
    \begin{center}
        \caption{\textbf{Image classification on datasets with high noise rate.} We compare to state-of-the-art approaches and report test top-1 accuracy ($\%$) on Clothing1M~\cite{xiao2015learning}
        (noise ratio $\sim38\%$). 
        We also show how much Jigsaw-ViT$+$NCT is above NCT with \textcolor{ForestGreen}{$\uparrow$}.}
        \scalebox{0.8}{ 
                \begin{tabular}{c|cccccc||ccc}
                \multirow{2}{*}{Method}   
                & \multirow{2}{*}{\begin{tabular}[c]{@{}c@{}}JO\\ {\cite{tanaka2018joint}}\end{tabular}} & \multirow{2}{*}{\begin{tabular}[c]{@{}c@{}}PLC\\ {\cite{zhang2021learning}}\end{tabular}} & \multirow{2}{*}{\begin{tabular}[c]{@{}c@{}}ELR+\\
                {\cite{liu2020early}}\end{tabular}} & \multirow{2}{*}{\begin{tabular}[c]{@{}c@{}}DivideMix\\
                {\cite{li2020dividemix}}\end{tabular}} & \multirow{2}{*}{\begin{tabular}[c]{@{}c@{}}S3\\ {\cite{feng2021s3}}\end{tabular}} & \multirow{2}{*}{\begin{tabular}[c]{@{}c@{}}NCT\\ {\cite{chen2021boosting,chen2022compressing}}\end{tabular}} & 
                \multicolumn{3}{c}{\bf Ours}\\
                &  &    &  &  &  &  
                & DeiT-Small/16~\cite{touvron2021training} & Jigsaw-ViT & Jigsaw-ViT$+$NCT \\ \hline
                Acc. (\%) & 72.2 & 74.0 & 74.8 & 74.8 & 74.9 & 75.0 
                & 71.6 & 72.4 & {\bf 75.4} (Comp.~NCT \textcolor{ForestGreen}{$\uparrow 0.4$})
                \\ \cdashline{1-10}
                Backbone                
                & \multicolumn{6}{c||}{ResNet-50~\cite{he2016deep}, \#params:  25.6M, FLOPS: 4G}
                & \multicolumn{3}{c}{DeiT-Small/16, \#params: 22M, FLOPS: 4.6G} 
                \\ \bottomrule[0.75pt]
                \end{tabular}}
        \label{tab::clothing1m}
    \end{center}
    \end{table*}
    
A more challenging classification problem is conducted to evaluate our Jigsaw-ViTs, that is, the image classification with noisy labels. 
Three popular real-world noisy label datasets are extensively evaluated: Animal-10N~\cite{song2019selfie}, Food-101N~\cite{lee2018cleannet} and Clothing1M~\cite{xiao2015learning}, where Animal-10N~\cite{song2019selfie} and Food-101N~\cite{lee2018cleannet} are with noisy labels at a relatively low ratio, Clothing1M~\cite{xiao2015learning} contains noisy labels at a high ratio.

\captionsetup[table*]{farskip=2pt,captionskip=1pt,aboveskip=4pt}
\begin{table*}[t]
    \centering
    \renewcommand{\arraystretch}{1.1}
    \caption{\textbf{Robustness to adversarial examples in black-box settings.} We report top-1 accuracy (\%) after attacks on ImageNet{-1K}~\cite{deng2009imagenet} validation set (higher numbers indicate better model robustness). 
    (a) Transfer-based attacks where adversarial examples are generated by attacking a surrogate model.
    (b) Query-based attacks where adversarial examples are generated by querying the target classifier for multiple times.
    We also show how much our Jigsaw-ViT model is above DeiT-Small/16~\cite{touvron2021training} with \textcolor{ForestGreen}{$\uparrow$}.}
        \begin{minipage}{\textwidth}
            \centering
            \scalebox{0.8}
            {\subfloat[Acc.~(\%) under transfer-based attacks.]{
            \begin{tabular}{c|c||lllll}
            Surrogate & Target & {\quad FGSM~\cite{goodfellow2014explaining}} 
            & BIM~\cite{kurakin2018adversarial}
            & PGD~\cite{madry2018towards} 
            & MI~\cite{dong2018boosting}
            & AA~\cite{croce2020reliable}
            \\ \hline
            \multirow{2}{*}{ViT-Small/16}   
            & DeiT-Small/16  
            & {\quad 32.8} 
            & 40.2 
            & 44.7 
            & 37.6 
            & 59.8 
            \\
            & Jigsaw-ViT 
            & {\quad \bf 34.8} \textcolor{ForestGreen}{$\uparrow 2.0$}
            & \bf 43.0 \textcolor{ForestGreen}{$\uparrow 2.8$}
            & \bf 47.7 \textcolor{ForestGreen}{$\uparrow 3.0$}
            & \bf 40.2 \textcolor{ForestGreen}{$\uparrow 2.6$}
            & \bf 62.5 \textcolor{ForestGreen}{$\uparrow 2.7$}
            \\ \cdashline{1-7}
            \multirow{2}{*}{ResNet-152} & DeiT-Small/16  
            & {\quad 59.0} & 65.9 & 67.9 & 65.3 
            & 70.6 
            \\
            & Jigsaw-ViT 
            & {\quad \bf 60.5} \textcolor{ForestGreen}{$\uparrow 1.5$} 
            & \bf 68.0 \textcolor{ForestGreen}{$\uparrow 2.1$} 
            & \bf 69.8 \textcolor{ForestGreen}{$\uparrow 1.9$}
            & \bf 66.8 \textcolor{ForestGreen}{$\uparrow 1.5$}
            & \bf 72.2 \textcolor{ForestGreen}{$\uparrow 1.6$}
            \\ \bottomrule[0.75pt]
            \end{tabular}
        \label{tab::transfer}}}
        \end{minipage}\hfill
        \begin{minipage}{\textwidth}
            \centering
            \scalebox{0.8}
            {\subfloat[Acc.~(\%) under query-based attacks.]{
            \begin{tabular}{c|c||cc}
            Attack & Num. queries & DeiT-Small/16 & Jigsaw-ViT 
            \\ \hline
            \multicolumn{1}{c|}{\multirow{4}{*}{{\begin{tabular}[c]{@{}c@{}}Square\\{\cite{andriushchenko2020square}}\end{tabular}}}} 
            & 50   & 49.3  
            &  \bf 52.1 \textcolor{ForestGreen}{$\uparrow 2.8$}      
            \\
            \multicolumn{1}{c|}{} & 100  & 36.6 
            & \bf 41.4 \textcolor{ForestGreen}{$\uparrow 4.8$} \\
            \multicolumn{1}{c|}{} & 200  & 22.8             
            & \bf 30.8 \textcolor{ForestGreen}{$\uparrow 8.0$} \\
            \multicolumn{1}{c|}{} & 500  & 7.2 
            & \bf 16.9 \textcolor{ForestGreen}{$\uparrow 9.7$}
            \\ \bottomrule[0.75pt]
            \end{tabular}
            \label{tab::square}}}
    \end{minipage}\hfill
    \label{tab::black-box}
\end{table*}

\captionsetup[table*]{farskip=2pt,captionskip=1pt,aboveskip=4pt}
\begin{table}[t]
    \centering
    \renewcommand{\arraystretch}{1.1}
    \caption{\textbf{Robustness to adversarial examples in white-box settings.} We report top-1 accuracy (\%) after attacks on ImageNet{-1K}~\cite{deng2009imagenet} validation set (higher numbers indicate better model robustness). 
    (a) White-box attacks with $L_{\infty}$-norm perturbation.
    (b) White-box attacks with $L_2$-norm perturbation.
    We also show how much our Jigsaw-ViT model is above DeiT-Small/16~\cite{touvron2021training} with \textcolor{ForestGreen}{$\uparrow$}.}
        \begin{minipage}{0.6\columnwidth}
            \centering
            \scalebox{0.8}
            {\subfloat[Acc.~(\%) under attacks with $L_{\infty}$-norm perturbation.]{
            \begin{tabular}{c|c||lll}
            Attack  & Model  & $\epsilon=4$ & $\epsilon=8$ & $\epsilon=16$ 
            \\ \hline
            \multirow{2}{*}{FGSM~\cite{goodfellow2014explaining}} & DeiT-Small/16 
            & 29.0  &  25.4 &  21.1  
            \\
            & Jigsaw-ViT    
            & {\bf 34.1} \textcolor{ForestGreen}{$\uparrow 5.1$}  
            & {\bf 30.2} \textcolor{ForestGreen}{$\uparrow 4.8$}  
            & {\bf 24.9} \textcolor{ForestGreen}{$\uparrow 3.8$}  
            \\ \cdashline{1-5}
            \multirow{2}{*}{PGD~\cite{madry2018towards}, Steps=5}  & DeiT-Small/16 
            & 1.5  & 0.3 & 0.1  \\
            & Jigsaw-ViT    
            & {\bf 5.3} \textcolor{ForestGreen}{$\uparrow 3.8$}  
            & {\bf 2.3} \textcolor{ForestGreen}{$\uparrow 2.0$}  
            & {\bf 1.3} \textcolor{ForestGreen}{$\uparrow 1.2$}  
            \\ \cdashline{1-5}
            \multirow{2}{*}{PGD~\cite{madry2018towards}, Steps=7}  & DeiT-Small/16 
            & 0.5 & 0.0 & 0.0  \\
            & Jigsaw-ViT    
            & {\bf 2.9} \textcolor{ForestGreen}{$\uparrow 2.4$}  
            & {\bf 0.6} \textcolor{ForestGreen}{$\uparrow 0.6$}  
            & {\bf 0.2} \textcolor{ForestGreen}{$\uparrow 0.2$}  
            \\ \bottomrule[0.75pt]
            \end{tabular}
        \label{tab::linf}}}
        \end{minipage}\hfill
        \begin{minipage}{0.4\textwidth}
            \centering
            \scalebox{0.8}
            {\subfloat[Acc.~(\%) under attacks with $L_2$-norm perturbation.]{
            \begin{tabular}{c|c||cc}
            Attack & Steps & DeiT-Small/16 & Jigsaw-ViT 
            \\ \hline
            \multicolumn{1}{c|}{\multirow{2}{*}{CW~\cite{carlini2017towards}}}
            & 10 & 10.0
            & {\bf 14.9} \textcolor{ForestGreen}{$\uparrow 4.9$}
            \\
            & 20 & 1.6
            & {\bf 4.7} \textcolor{ForestGreen}{$\uparrow 3.1$}
            \\ \bottomrule[0.75pt]
            \end{tabular}
            \label{tab::l2}}}
    \end{minipage}\hfill
    \label{tab::white-box}
\end{table}

\paragraph{Training details} 
Animal-10N~\cite{song2019selfie} consists of 10 classes of animal images crawled online with manually annotated labels. 
The dataset consists of $50{\small,}000$ training images with label noise ratio $\sim$8\% and $5{\small,}000$ clean testing images. 
Food-101N~\cite{lee2018cleannet} contains $310{\small,}009$ training images of food recipes collected online and are classified 101 classes with noise ratio $\sim$20\%. 
Following~\cite{lee2018cleannet}, the learnt models should be evaluated on the test set of Food-101 of $25{\small,}250$ clean labeled images. 
Clothing1M~\cite{xiao2015learning} is a large-scale dataset containing 1 million images of clothing crawled online. 
The dataset is categorized into 14 classes, containing $1{\small,}000{\small,}000$ training images with noise ratio $\sim$38\% and $10{\small,}526$ test images. 
We follow the preprocessing in~\cite{zhang2021learning,chen2022compressing} for this dataset.

We train all the models from scratch, unlike most methods requiring extra data and learning from ImageNet{-1K} pretrained models \cite{liu2020early,li2020dividemix,zhang2021learning,feng2021s3}, e.g.,~methods in Table~\ref{tab::food101} and Table~\ref{tab::clothing1m} use ImageNet{-1K} pretrained ResNet-50~\cite{he2016deep}.
We use AdamW with a weight decay of 0.05 and train for 400K iterations with batch size 128. 
The training starts with a linear learning rate warm-up for 20{K} steps and cosine learning rate decay with a maximum learning rate of 1e-3 and a minimum of 1e-6.

\paragraph{Results} 
We report top-1 accuracy on datasets with low noise rate, i.e., Animal-10N~\cite{song2019selfie} and Food-101N~\cite{lee2018cleannet} in Table~\ref{tab::lownoiserate}. 
Interestingly, the baseline DeiT-Small/16~\cite{touvron2021training} achieves promising results on both datasets and already outperforms some competitive approaches, which demonstrates the powerful capabilities of ViTs on this task. 
Note that, in terms of model complexity, this ViT is comparable to ResNet-50 and much lighter than VGG architectures which are commonly used in the community.
Our Jigsaw-ViT trained under $\mathcal{L}_{\text{total}}$ consistently outperforms DeiT-Small/16 trained under a single $\mathcal{L}_{\text{cls}}$ with substantial improvements.
Moreover, our methods also achieve the best performances on both datasets in Table~\ref{tab::lownoiserate} among all compared state-of-the-art methods specifically designed for this noisy label task.
These results indicate that our deployed auxiliary loss can implicitly serve as a practical regularization for learning with noisy labels. 
Similarly in Table~\ref{tab::clothing1m}, we observe the same tendency on the experiments with Clothing1M~\cite{xiao2015learning} dataset.
Despite the high noisy ratio on Clothing1M, our Jigsaw-ViTs maintain to achieve promising results.

Additionally, it is worth mentioning that our method can serve as complementary strategies to other state-of-the-art methods to further boost their performances.
In particular, we incorporate our Jigsaw-ViT to NCT~\cite{chen2021boosting} ({Jigsaw-ViT+NCT}).
NCT is a two-stage method designed for combating label noise. 
Notably, by implementing NCT with our Jigsaw-ViT, the top-1 accuracy of NCT is improved by 0.4\%, achieving the state-of-the-art result of 75.4\%.
These experiments not only verify the superiority of Jigsaw-ViTs as a stand-alone method, but also the effectiveness as a promising complementary tool to existing methods in combating label noise with boosted performances.

\subsection{Robustness to adversarial examples}
\label{subsec::adv}
In this subsection, we investigate the robustness of Jigsaw-ViT against adversarial examples with perturbations on input images under both black-box and white-box attaches.

\paragraph{Training details}
Adversarial examples are crafted images by adding visually imperceptible perturbations to the clean images, which deteriorates the model predictions. 
In accordance to Section~\ref{subsec:imagenet}, we conduct the experiments on the models trained with ImageNet{-1K}~\cite{deng2009imagenet}.
Following~\cite{xie2019improving}, the validation set with $50{\small,}000$ images are used as clean samples to generate adversarial examples. 
In the experiments, both black-box and white-box settings are considered for adversarial attacks. 

In black-box settings, we first consider the transfer-based attacks where adversarial examples are generated by attacking  surrogate  models and then fed into the target models (e.g., our Jigsaw-ViTs) to evaluate the robustness performance. 
We then consider query-based attacks, which only require multiple queries of the outputs of the target models to perform the attacks. 
Following the settings in \cite{xie2019improving,wu2020skip}, methods used for generating the adversarial examples in transfer-based attacks are: FGSM~\cite{goodfellow2014explaining}, 
basic iterative method (BIM)~\cite{kurakin2018adversarial}, 
PGD~\cite{madry2018towards},
momentum iterative boosting (MI)~\cite{dong2018boosting} and the ensemble AA~\cite{croce2020reliable} which include white-box attacks.
These attacks are crafted under maximum $L_{\infty}$-norm perturbation $\epsilon=16$ with respect to pixel values in $[0,255]$, step size $\alpha=2$, and number of steps $10$ if it is a multi-step attack such as BIM, PGD and MI~\cite{wu2020skip}.
We adopt the pretrained ViT-Small/16 in~\cite{dosovitskiy2021an} and pretrained ResNet-152~\cite{he2016deep} on ImageNet{-1K}~\cite{deng2009imagenet} as the surrogate models which are white-box with gradient available.
The target victim models are DeiT-Small/16 and our Jigsaw-ViT whose gradients are inaccessible.
For query-based attacks, we consider the popular square attack (Square)~\cite{andriushchenko2020square} with $L_\infty$-norm perturbation $\epsilon=16$ and  different querying numbers  $\{50, 100, 200, 500\}$.

In white-box settings, the adversarial examples are generated by directly attacking the accessible target victim models (DeiT-Small/16 and our Jigsaw-ViT).
We consider both gradient-based attacks including FGSM and PGD, and one typical non-gradient-based attack named CW~\cite{carlini2017towards}.
As in \cite{wu2020skip}, we consider FGSM and PGD with the commonly-used $L_{\infty}$-norm bounds of perturbations as $\epsilon\in[4,8,16]$.
For CW, the $L_2$-norm perturbations are used with box-constraint parameter $c=1$ and step size varying in $\{10,20\}$.

\paragraph{Results} 
Table~\ref{tab::black-box} and Table~\ref{tab::white-box} report
the top-1 accuracy ($\%$) of the victim DeiT-Small/16 and our Jigsaw-ViT against black-box and white-box attacks, respectively. 
For adversarial examples generated by various attacks on different surrogate models in Table~\ref{tab::black-box}, the performances of target models all degrade drastically compared to their clean counterparts in Table~\ref{tab::imagenet}, demonstrating the challenge of this task.
In contrast, our Jigsaw-ViTs provides distinctively higher accuracy than standard ViTs under both transfer-based attacks and query-based ones.
Such robustness of Jigsaw-ViT over DeiT-Small/16 becomes even more significant as the square attack utilizes more queries for stronger crafts in Table~\ref{tab::square}.
Specifically, Jigsaw-ViT exceeds DeiT-Small/16 by $9.7\%$ with maximal 500 queries, compared the $2.8\%$ improvement with maximal 50 queries.

In the white-box attacking results in Table~\ref{tab::white-box}, in addition to outperforming DeiT-Small/16 in all cases, our Jigsaw-ViT has non-zero accuracy in cases where DeiT-Small/16 is completely crafted by the PGD attack ($\text{steps}=7$, $\epsilon\in\{8,16\}$) in Table~\ref{tab::linf}. 
Results in Table~\ref{tab::l2} relating to non-gradient-based CW attacks further demonstrate the effectiveness of the proposed Jigsaw-ViT, together verifying that injecting the proposed jigsaw puzzle flow to ViTs successfully provides improvements on the robustness against adversarial examples under various settings.

\subsection{Ablation study}
\label{sec::ablation}
In this subsection, we first study the effect of positional embeddings in the jigsaw branch, and then investigate the impacts of the two hyper-parameters in Jigsaw-ViT: the loss balancing coefficient $\eta$ in \eqref{eq::obj_func}, the mask ratio $\gamma$ of jigsaw image patches. 

\paragraph{Effect of positional embedding in the jigsaw branch} 
We conduct experiments on ImageNet{-1K}~\cite{deng2009imagenet} and noisy label datasets including Animal-10N~\cite{song2019selfie}, Food-101N~\cite{lee2018cleannet} and Clothing1M~\cite{xiao2015learning}. 
Results are given in Table~\ref{tab::pos-emb}, showing that the removal of positional embeddings in the jigsaw branch helps provide consistent improvement over all tested datasets, which validates our effective design of the jigsaw branch in ViTs.

\paragraph{Impact of $\eta$ and $\gamma$} 
The experimental investigations on the two hyper-parameter involved in the proposed Jigsaw-ViTs are conducted on datasets Animal-10N~\cite{song2019selfie}, Food-101N~\cite{lee2018cleannet} and Clothing1M~\cite{xiao2015learning}. 
The results are illustrated in Table~\ref{tab::ablation}. 
First, Jigsaw-ViT provides consistent improvement on all three datasets compared to its non-jigsaw counterpart i.e., DeiT-Small/16 ($\eta=0$). Moreover, the improvement is quite robust to the choices of both $\eta$ and $\gamma$. 
Second, injecting the jigsaw puzzle solving with non-zero mask ratio to ViT indeed brings consistent performance boost over standard ViT, as we still observe that non-zero mask ratio $\gamma$ shows better performances than the case of $\gamma=0$ on different datasets and lower mask ratios can already lead to good improvements.
Hence, these evidences all demonstrate the effectiveness of the proposed approach. 

\captionsetup[table]{farskip=2pt,captionskip=1pt,aboveskip=4pt}
    \begin{table}[t]
    \renewcommand{\arraystretch}{1.1}
    \begin{center}
        \caption{\textbf{Comparisons between Jigsaw-ViT w/ and w/o pos.~emb.}
        We report test top-1 accuracy (\%) on ImageNet{-1K}~\cite{deng2009imagenet} and noisy label datasets including Animal-10N~\cite{song2019selfie}, Food-101N~\cite{lee2018cleannet} and Clothing1M~\cite{xiao2015learning}.
        We also show how much Jigsaw-ViT w/o pos.~emb. is above Jigsaw-ViT w/ pos.~emb.~with \textcolor{ForestGreen}{$\uparrow$}.}
        \scalebox{0.8}{ 
            \begin{tabular}{c||llll}
            \multirow{2}{*}{} & \multirow{2}{*}{ImageNet{-1K}} 
            & Animal-10N  & Food-101N &  Clothing1M 
            \\
            &  
            & \multicolumn{1}{c}{Noise $\sim 8\%$} 
            & \multicolumn{1}{c}{Noise $\sim 20\%$} 
            & \multicolumn{1}{c}{Noise $\sim 38\%$} 
            \\ \hline
            w/ pos. emb.      
            & 80.3 & {\quad 87.3} & {\quad 84.7} & {\quad 71.1}
            \\ 
            w/o pos. emb.     
            & {\bf 80.5} \textcolor{ForestGreen}{$\uparrow 0.2$}     
            & {\quad \bf 88.7} \textcolor{ForestGreen}{$\uparrow 1.4$}   
            & {\quad \bf 86.5} \textcolor{ForestGreen}{$\uparrow 1.8$}     
            & {\quad \bf 72.4} \textcolor{ForestGreen}{$\uparrow 1.3$}     
            \\ \bottomrule[0.75pt]
            \end{tabular}}
        \label{tab::pos-emb}
    \end{center}
    \end{table}
    
    \captionsetup[table]{farskip=2pt,captionskip=1pt,aboveskip=4pt}
    \begin{table}[t]
        \centering
        \renewcommand{\arraystretch}{1.1}
        \caption{{\bf Ablation study on real-world noisy label datasets.} We report test top-1 accuracy ($\%$) on the following datasets: 
        Animal-10N~\cite{song2019selfie} (noise ratio $\sim8\%$), 
        Food-101N~\cite{lee2018cleannet} (noise ratio $\sim20\%$) and 
        Clothing1M~\cite{xiao2015learning} (noise ratio $\sim38\%$). Note that $\eta=0$ corresponds to DeiT-Small/16~\cite{touvron2021training}.}
        \begin{minipage}{\columnwidth}
            \centering
            \scalebox{0.8}{
                \begin{tabular}{c|cc||ccc}
                \multirow{2}{*}{Method} &
                \multirow{2}{*}{$\eta$} & Mask & Animal-10N
                & Food-101N
                & Clothing1M \\ 
                & & ratio
                & Noise $\sim8\%$
                & Noise $\sim20\%$
                & Noise $\sim38\%$
                \\ 
                \hline
                DeiT-Small/16 & 0  & --\-  & 87.2 & 84.2 & 71.6  \\ \hline
                \multirow{12}{*}{Jigsaw-ViT} & \multirow{4}{*}{0.5}    & 0 & 88.6 & 86.1& 71.9 \\
                & & 0.1  & 88.7 &  85.8 & 71.9       \\
                & & 0.2   & 88.7  & 86.1  & 71.8          \\ 
                & & {0.5} & {88.1} & {86.2} & {71.8}  \\ 
                \cdashline{2-6}
                & \multirow{4}{*}{1}  & 0 &  88.6& 86.1& 72.3 \\
                & & 0.1   & 88.6 & 86.4& 72.2   \\
                & & 0.2   & 88.7 & 86.3& 72.1        \\ 
                & & {0.5} & {88.3} & {\bf{86.7}} & {71.5}         \\ 
                \cdashline{2-6}
                & \multirow{4}{*}{2} & 0  & 88.6 &  86.4& 71.7  \\
                & & 0.1 & 88.7&  86.4& \bf 72.4        \\
                & & 0.2    & \bf 89.0 & 86.5 & 71.9        \\
                & & {0.5} & {87.8} & {86.1} &  {71.6}    \\ 
                \bottomrule[0.75pt]
                \end{tabular}}
        \end{minipage}\hfill
        \label{tab::ablation}
    \end{table}

    \captionsetup[table]{farskip=2pt,captionskip=1pt,aboveskip=4pt}
    \begin{table}[t]
    \renewcommand{\arraystretch}{1.1}
    \begin{center}
        \caption{{\textbf{Jigsaw puzzle solving as a pretext task.}
        We report test top-1 accuracy (\%) of models trained with jigsaw puzzle solving being the pretext task (entry ``Jigsaw") and without it (entries ``\xmark") on noisy label datasets.
        We also show how much DeiT with jigsaw puzzle solving as pretext task or as auxiliary loss improves over DeiT with \textcolor{ForestGreen}{$\uparrow$}.}}
        \scalebox{0.8}{ 
            \begin{tabular}{c|c||lll}
            {Method} 
            & \begin{tabular}[c]{@{}c@{}}{Pretext}\\ {task}\end{tabular}
            & \begin{tabular}[c]{@{}c@{}}{Animal-10N}\\ {Noise $\sim8\%$}\end{tabular} 
            & \begin{tabular}[c]{@{}c@{}}{Food-101N}\\ {Noise $\sim20\%$}\end{tabular} 
            & \begin{tabular}[c]{@{}c@{}}{Clothing1M}\\ {Noise $\sim38\%$}\end{tabular} 
            \\ \hline
            \multirow{2}{*}{{DeiT-Small/16}} & {\xmark}  & {87.2} & {84.2}  & {71.6}       
            \\
            & {Jigsaw} 
            & {88.0} \textcolor{ForestGreen}{$\uparrow 0.8$} 
            & {\bf{87.3}} \textcolor{ForestGreen}{$\uparrow 3.1$} 
            & {71.9} \textcolor{ForestGreen}{$\uparrow 0.3$} 
            \\ 
            \cdashline{1-5}
            {Jigsaw-ViT} & {\xmark}  
            & {\bf{89.0} \textcolor{ForestGreen}{$\uparrow 1.8$}} 
            & {86.7 \textcolor{ForestGreen}{$\uparrow 2.5$}}
            & {\bf{72.4} \textcolor{ForestGreen}{$\uparrow 0.8$}}  
            \\ 
            \bottomrule[0.75pt]
        \end{tabular}}
        \label{tab::pretext}
    \end{center}
    \end{table}

    \captionsetup[table]{farskip=2pt,captionskip=1pt,aboveskip=4pt}
    \begin{table}[t]
    \renewcommand{\arraystretch}{1.1}
    \begin{center}
        \caption{{\textbf{Semantic segmentation as a downstream task.}
        Performance of Segmenters~\cite{strudel2021segmenter} on ADE20K~\cite{zhou2019semantic} validation set with pretrained weights provided by DeiT~\cite{touvron2021training} and Jigsaw-ViT trained on ImageNet{-1K}~\cite{deng2009imagenet}.
        We also show how much Segmenter with Jigsaw-ViT pretrained weight improves over its DeiT pretrained counterpart with \textcolor{ForestGreen}{$\uparrow$}.}}
        \scalebox{0.8}{ 
        \begin{tabular}{c|c||ll}
           {Method}  & {Pretrained}  & {Pixel Acc.} & {mIoU (SS)} 
            \\ \hline
            \multirow{2}{*}{{Seg-Small/16}} 
            & {DeiT-Small/16} &  {80.2} & {42.3} 
            \\
            & {Jigsaw-ViT-Small}
            & {{\bf 80.6}} \textcolor{ForestGreen}{$\uparrow 0.4$} 
            & {{\bf 42.9}} \textcolor{ForestGreen}{$\uparrow 0.6$}          
            \\ \cdashline{1-4}
            \multirow{2}{*}{{Seg-Base/16}} & {DeiT-Base/16}    
            & {81.1}   
            & {45.1}    
            \\
            & {Jigsaw-ViT-Base} 
            & {\bf 81.3} \textcolor{ForestGreen}{$\uparrow 0.2$}       
            & {\bf 45.7} \textcolor{ForestGreen}{$\uparrow 0.6$}         
            \\ \bottomrule[0.75pt]
            \end{tabular}}
        \label{tab::seg}
    \end{center}
    \end{table}

\subsection{{Further discussions: pretext and downstream tasks}}
\label{sec::further}
To further explore potentials of Jigsaw-ViT,
we investigate the settings of both pretext and downstream tasks.
First, we set the jigsaw puzzle solving as a pretext task so as to testify the necessity of building it as an auxiliary loss term.
Second, we consider whether Jigsaw-ViT can benefit downstream tasks, such as semantic segmentation.

\paragraph{{Jigsaw puzzle solving as a pretext task}}
We set jigsaw puzzle solving, which is a self-supervised learning problem, as a pretext task for learning with noisy labels.
During the pretext training, we adopt AdamW with a weight decay of 0.05 and train for 200K iterations with batch size 128.
We set a linear learning rate warm-up for 20K steps and cosine learning rate decay with a maximum learning rate of 1e-3 and a minimum of 1e-6.
Then, the backbone is fine-tuned on the same noisy label dataset with only $\mathcal{L}_{\text{cls}}$ used.
Table~\ref{tab::pretext} gives the comparisons of models trained with (entry ``Jigsaw") and without (entry ``\xmark") the jigsaw pretext task.
Using jigsaw puzzle solving as a pretext task successfully outperforms training DeiTs from scratch without it.
It shows that building jigsaw puzzle solving as an auxiliary loss term, i.e.,~Jigsaw-ViT, commonly leads to better performances than 
setting it as a pretext task.
Rather than utilizing jigsaw puzzle solving as a pretext task, building it as an auxiliary loss term in the end-to-end training can in general be encouraged in usage to enhance model's robustness against noisy labels.

\paragraph{{Semantic segmentation as a downstream task}}
{Semantic segmentation aims to label each image pixel with a corresponding category of the underlying object.
Segmenter~\cite{strudel2021segmenter} builds upon ViTs and extends to semantic segmentation with extra semantic class tokens and a mask transformer decoder.
Since this SOTA should be implemented on a pretrained ViT, we set the weights of DeiT and Jigsaw-ViT in Table~\ref{tab::imagenet} as pretrained weights for training on the challenging ADE20K dataset~\cite{zhou2019semantic}. 
Table~\ref{tab::seg} shows that Segmenters with Jigsaw-ViT pretrained weights outperform their DeiT pretrained counterparts with respect to both mIoU with single scale (SS) inference, and pixel accuracy.
These experiments verify that Jigsaw-ViT is also promising to benefit downstream task such as semantic segmentation, further showing the potentials of our proposed method.}

\section{Conclusion}
\label{sec:conclusion}
In this paper, we inject jigsaw flow into the standard ViT by solving classical jigsaw puzzle as a self-supervised auxiliary loss during optimization, namely Jigsaw-ViT.
To better utilize jigsaw puzzle solving, we propose to discard the positional embeddings in the jigsaw flow to avoid being directly hinted about the explicit position clues. 
Meanwhile, we observe that dropping some patches, 
i.e.,~patch masking, also plays a positive role in this procedure. 
The introduced auxiliary loss by injecting the jigsaw flow with the aforementioned two strategies not only enhances ViTs' performances for large-scale classification on ImageNet, but also attains stronger resistance to label noise and adversarial examples.
These show that Jigsaw-ViTs improve standard ViTs in both generalization and robustness, even though such two aspects are usually considered as a trade-off.
Our proposed method is an easy-to-reproduce, and yet a very effective tool to boost the performance of ViTs and even a plug-in complementary tool to enhance other methods.

\section*{Acknowledgements}
This work was jointly supported by the European Research Council under the European Union’s Horizon 2020 research and innovation program/ERC Advanced Grant E-DUALITY (787960), Research Council KU Leuven: Optimization framework for deep kernel machines C14/18/068, The Research Foundation–Flanders (FWO) projects: GOA4917N (Deep Restricted kernel Machines: Methods and Foundations), Ph.D./Postdoctoral grant, the Flemish Government (AI Research Program), EU H2020 ICT-48 Network TAILOR (Foundations of Trustworthy AI-Integrating Reasoning, Learning and Optimization), Leuven.AI Institute.

\bibliographystyle{unsrt}  
\bibliography{egbib}

\section*{Supplementary Material}

\section*{1\quad Ablation study on ImageNet}
In this section, we provide the ablation study with $\eta$ which balances the two losses, and $\gamma \in [0,1)$ which is the mask ratio of jigsaw image patches.
We train both models (DeiT-Small/16 and our Jigsaw-ViT) from scratch following the same training protocols as illustrated in the main paper where AdamW is used as the optimizer with a weight decay of 0.05, the overall batch size is set to 1,024, and 300 training epochs. 
The base learning rate is 5e-4. 

As shown in Table~\ref{tab::supp-imagenet}, when we choose the proper configuration, i.e.~$\eta=0.1$ and $\gamma=0.5$, Jigsaw-ViT can significantly boost the test accuracy of the standard DeiT-Small/16 by 0.7\%.
Besides, there are two insights that can be seen from the results:
\textit{i)}
adding our jigsaw loss consistently improves the results over the baseline DeiT-Small/16 in most settings;
\textit{ii)}
Our design of masking part of patches ($\gamma>0$) outperforms reordering all the patches ($\gamma=0$).
These experimental results verify the effectiveness of our proposed method.

Additionally, we consider Swin transformers~[9] as larger and more diverse backbones.
In Swin transformer, image patches are merged from small-sized patches in the initial layer to large-sized patches in the output layer.
Our jigsaw branch is set to predict the absolute positions of the final merged patches.
We follow the same training protocol of pretraining on ImageNet-22K and fine-tuning on ImageNet-1K in~[9]. 
To be specific, we compare the fine-tuning performance initialized with the ImageNet-22K pretrained weights provided in~[9] and then use the cross entropy loss (baseline) and our method to further proceed the optimization, where Jigsaw-ViT is simply set with $\eta=0.1$, $\gamma=0$ for all cases.

Results of setting Swin as backbone for Jigsaw-ViTs on ImageNet are given in Table~\ref{tab::supp-swin}.
For almost all architectures on both datasets, adding jigsaw branch consistently improves upon the baselines by fine-tuning, verifying the effectiveness of our Jigsaw-ViT in attaining better generalization performances on large-scale image classification. 
Our jigsaw solving enables to fine-tune Swin-Base from 82.6\% to 85.0\%, though it slightly underperforms the fine-tuning result by the cross entropy loss. 
Note that the design of local windows in Swin implicitly leaks positional information, which might lead to inconsistent improvement over different settings.

\captionsetup[table]{farskip=2pt,captionskip=1pt,aboveskip=4pt}
\begin{table*}[t]
    \centering
    \renewcommand{\arraystretch}{1.1}
    \caption{\textbf{Image classification on ImageNet validation set.} We compare to DeiT-Small/16 and report top-1 accuracy ($\%$). We also show how much each Jigsaw-ViT model is above or below DeiT-Small/16, which are marked with \textcolor{ForestGreen}{$\uparrow$} and \textcolor{RedOrange}{$\downarrow$} respectively. }
        \scalebox{0.8}{
            \begin{tabular}{c||c|ccccc}
            &
            $\quad\eta$ $\backslash$ $\gamma\quad$ &
            $\quad\gamma=0\quad$ & $\quad\gamma=0.1\quad$ & $\quad\gamma=0.2\quad$ &
            $\quad\gamma=0.5\quad$ & $\quad\gamma=0.8\quad$ \\ \hline 
            $\quad$DeiT-Small/16 $\quad$ &
            $\quad\eta=0\quad$ & $\quad79.8\quad$ & $\quad-\quad$ & $\quad-\quad$ &
            $\quad-\quad$ & $\quad-\quad$ 
            \\ \hline
            \multirow{6}{*}{Jigsaw-ViT (Ours)} &
            $\eta=0.1$& 80.3  & 80.3 & \bf 80.5 & \bf 80.5 & 80.4 \\
            &Comp. DeiT-Small/16&  \textcolor{ForestGreen}{$\uparrow 0.5$}   & \textcolor{ForestGreen}{$\uparrow 0.5$}
            & \textcolor{ForestGreen}{$\uparrow 0.7$} 
            & \textcolor{ForestGreen}{$\uparrow 0.7$}  
            & \textcolor{ForestGreen}{$\uparrow 0.6$}  \\ \cdashline{2-7}
            & $\eta=0.5$& 80.2  & 80.2 & 80.1 & 80.2  & 80.3 \\
            &Comp. DeiT-Small/16&  \textcolor{ForestGreen}{$\uparrow$ 0.4}   & \textcolor{ForestGreen}{$\uparrow 0.4$}
            & \textcolor{ForestGreen}{$\uparrow 0.3$} 
            & \textcolor{ForestGreen}{$\uparrow 0.4$}  
            & \textcolor{ForestGreen}{$\uparrow 0.5$}  \\ \cdashline{2-7}
            & $\eta=1.0$& 80.0  & 80.0 & 79.8 &  79.5 &  80.3 \\
            &Comp. DeiT-Small/16&  \textcolor{ForestGreen}{$\uparrow$ 0.2}   & \textcolor{ForestGreen}{$\uparrow 0.2$}
            & \textcolor{ForestGreen}{$\uparrow 0$} 
            & \textcolor{RedOrange}{$\downarrow 0.3$}  
            & \textcolor{ForestGreen}{$\uparrow 0.5$}  \\ 
            \bottomrule[0.75pt]
        \end{tabular}}
    \label{tab::supp-imagenet}
\end{table*}

\captionsetup[table]{farskip=2pt,captionskip=1pt,aboveskip=4pt}
    \begin{table}[t]
        \centering
        \renewcommand{\arraystretch}{1.1}
        \caption{{\textbf{Image classification on ImageNet-1K~\cite{deng2009imagenet} validation set and ImageNet V2~\cite{recht2019imagenet}.} 
        We compare to Swin~\cite{liu2021swin} with different capacities and report top-1 accuracy ($\%$).
        We also show how much each Jigsaw-ViT model is above or below the baseline, which are marked with \textcolor{ForestGreen}{$\uparrow$} and \textcolor{RedOrange}{$\downarrow$} respectively.}}
        \scalebox{0.8}{
            \begin{tabular}{c|c||cc|cl}
            \multirow{2}{*}{{Backbone}} 
            & \multirow{2}{*}{{\#params}}
            & \multicolumn{2}{c|}{{ImageNet{-1K}}} & \multicolumn{2}{c}{{ImageNet V2}} 
            \\
            & & {Baseline} & {Jigsaw-ViT} & {Baseline} & {Jigsaw-ViT}         
            \\ \hline
            {Swin-Tiny} 
            & {28M}
            & {80.9} 
            & {\bf 81.2 \textcolor{ForestGreen}{$\uparrow 0.3$}} 
            & {70.5}
            & {\bf 70.7 \textcolor{ForestGreen}{$\uparrow 0.2$}}
            \\
            {Swin-Small} 
            & {50M}
            & {83.2} 
            & {\bf{83.6} \textcolor{ForestGreen}{$\uparrow 0.4$}} 
            & {73.6}  
            & {\bf{73.8} \textcolor{ForestGreen}{$\uparrow 0.2$}} 
            \\
            {Swin-Base} 
            & {88M}
            & {\bf 85.2} 
            & {85.0 \textcolor{RedOrange}{$\downarrow 0.2$}}
            & {75.3} 
           & {\bf 75.4 \textcolor{ForestGreen}{$\uparrow 0.1$}}       
            \\ \bottomrule[0.75pt]
            \end{tabular}}
        \label{tab::supp-swin}
\end{table}

\section*{2\quad Visualizing attention maps}
We show more examples of the visualization maps of  self-attention module with DeiT-Small/16 and Jigsaw-ViT trained on ImageNet. 
We base our visualization on the official codes of DINO\footnote{\label{note1}\url{https://github.com/facebookresearch/dino}}.
Note that the original code of DINO provides attention map for each self-attention head, leading to 6 attention maps for one input image when using DeiT-Small/16 as backbone.
For fair and better comparison, we concatenate all heads as one so as to provide only one attention map for each input image.

\begin{figure*}[ht]
		\centering
		\begin{minipage}{0.1\textwidth}
		\centering
		\footnotesize{Input}
		\end{minipage}\hfill
		\begin{minipage}{0.9\textwidth}
		\centering
		\includegraphics[width=0.1\columnwidth,height=0.1\columnwidth]{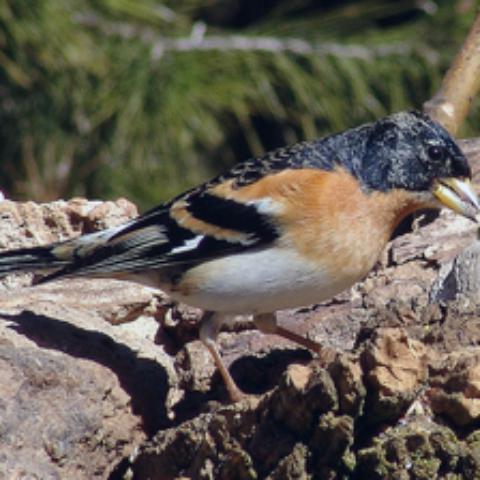}
		\includegraphics[width=0.1\columnwidth,height=0.1\columnwidth]{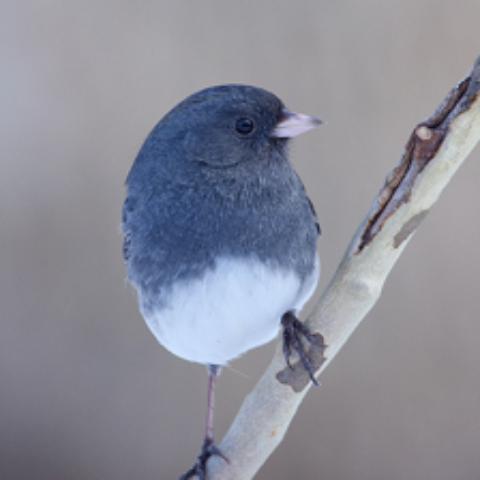}
		\includegraphics[width=0.1\columnwidth,height=0.1\columnwidth]{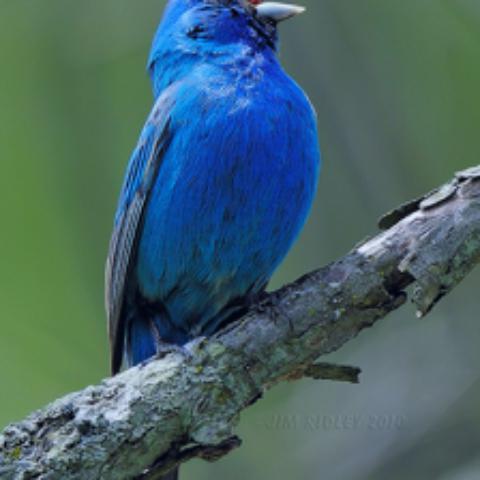}
		\includegraphics[width=0.1\columnwidth,height=0.1\columnwidth]{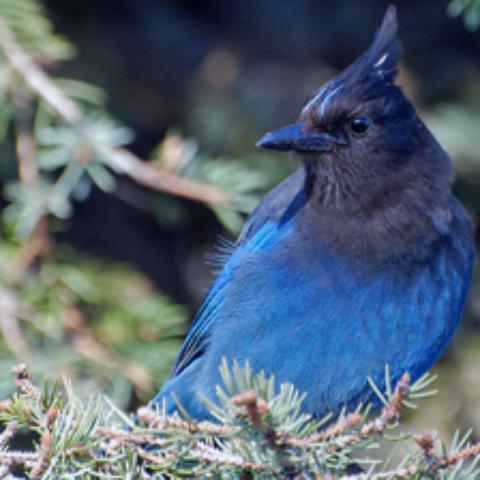}
		\includegraphics[width=0.1\columnwidth,height=0.1\columnwidth]{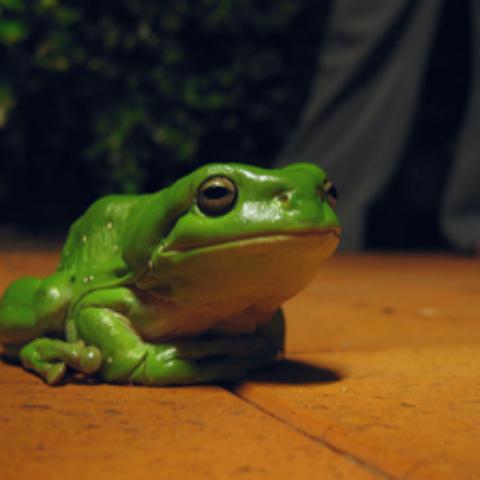}
		\includegraphics[width=0.1\columnwidth,height=0.1\columnwidth]{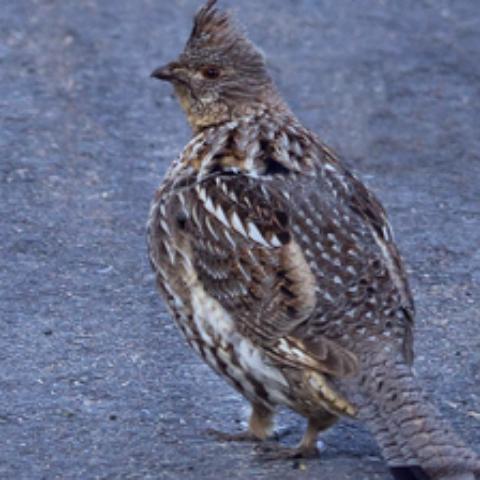}
		\includegraphics[width=0.1\columnwidth,height=0.1\columnwidth]{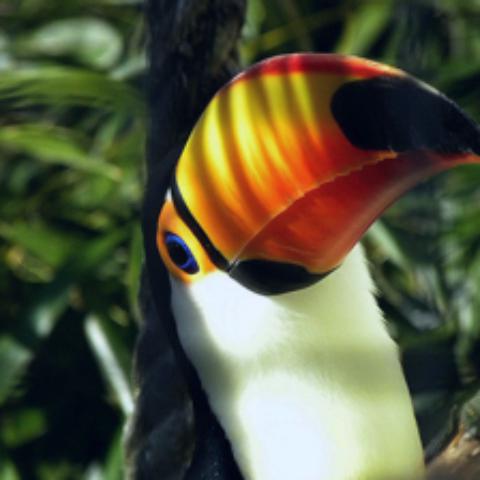}
		\includegraphics[width=0.1\columnwidth,height=0.1\columnwidth]{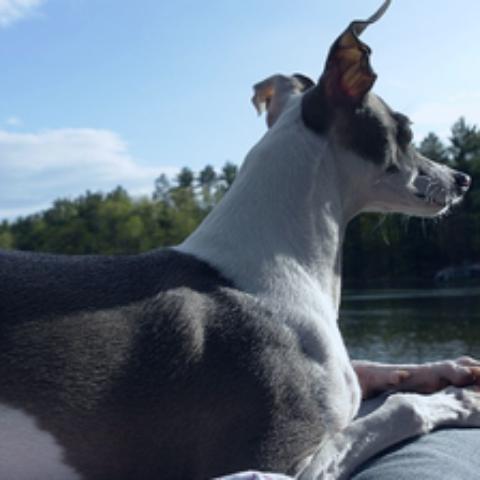}
		\includegraphics[width=0.1\columnwidth,height=0.1\columnwidth]{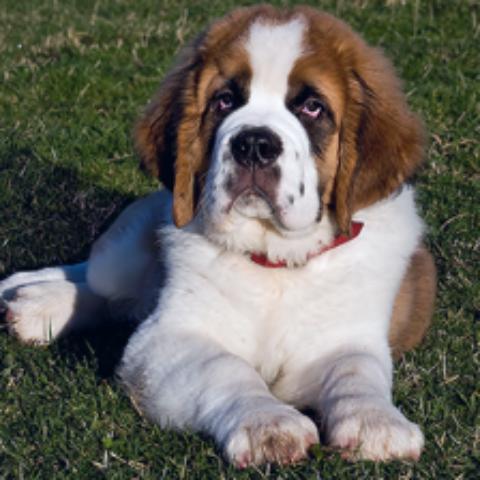}
		\end{minipage}\hfill
		\begin{minipage}{0.1\textwidth}
		\centering
		\footnotesize{DeiT-Small/16}
		\end{minipage}\hfill
		\begin{minipage}{0.9\textwidth}
		\centering
		\includegraphics[width=0.1\columnwidth,height=0.1\columnwidth]{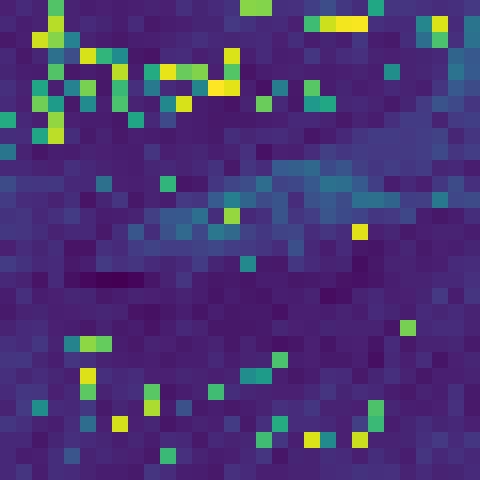}
		\includegraphics[width=0.1\columnwidth,height=0.1\columnwidth]{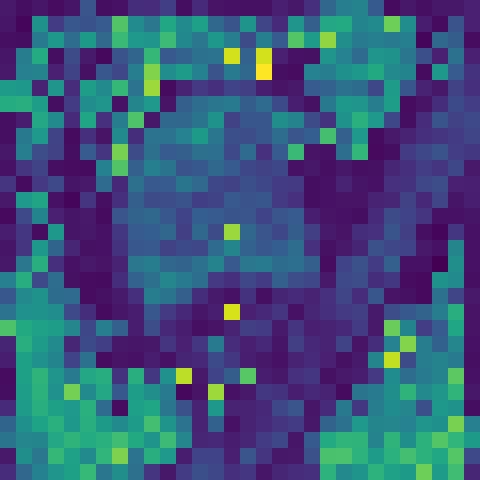}
		\includegraphics[width=0.1\columnwidth,height=0.1\columnwidth]{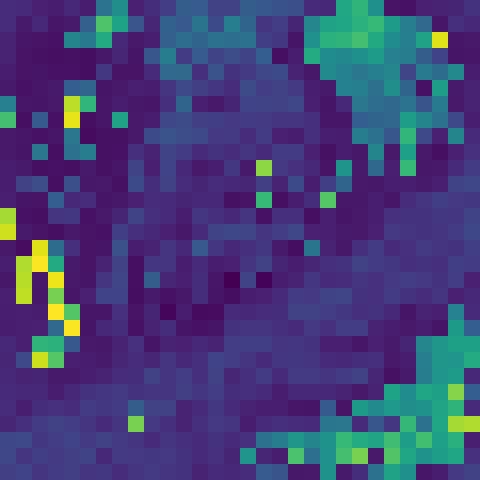}
		\includegraphics[width=0.1\columnwidth,height=0.1\columnwidth]{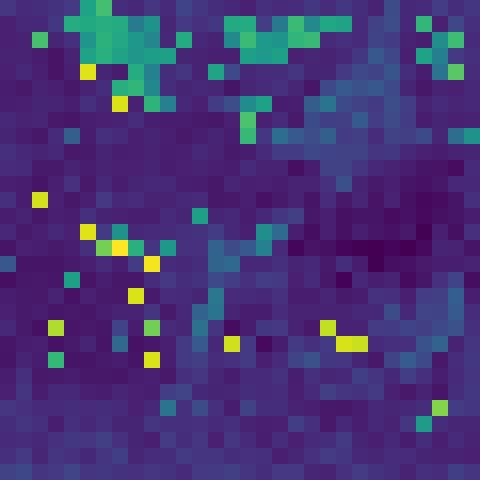}
		\includegraphics[width=0.1\columnwidth,height=0.1\columnwidth]{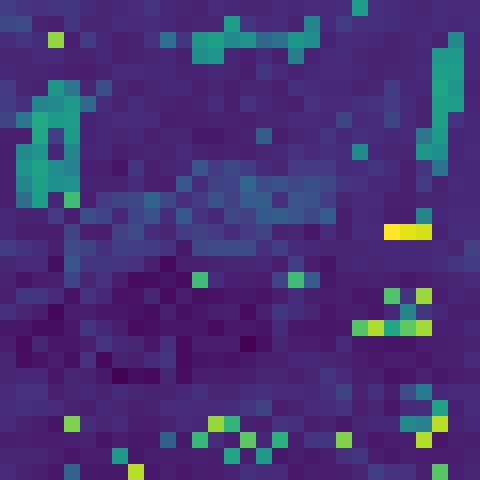}
		\includegraphics[width=0.1\columnwidth,height=0.1\columnwidth]{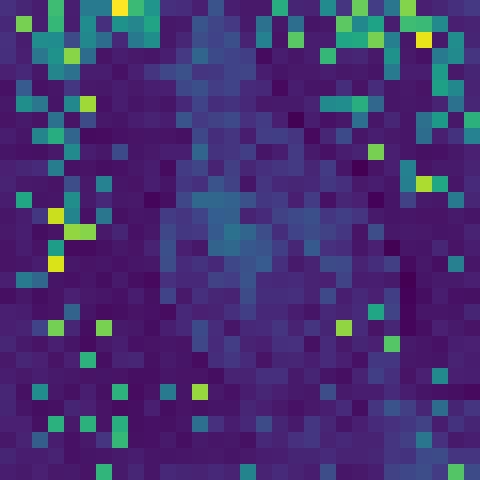}
		\includegraphics[width=0.1\columnwidth,height=0.1\columnwidth]{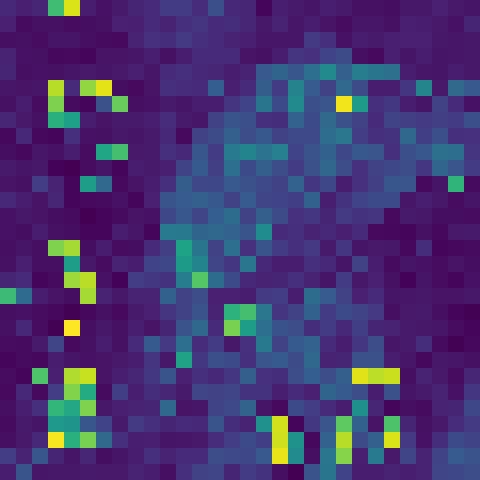}
		\includegraphics[width=0.1\columnwidth,height=0.1\columnwidth]{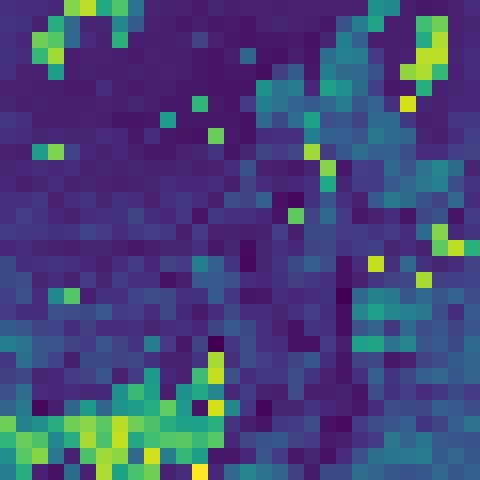}
		\includegraphics[width=0.1\columnwidth,height=0.1\columnwidth]{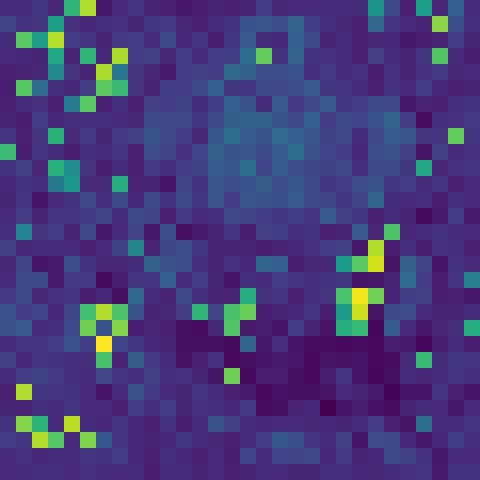}
		\end{minipage}\hfill
		\begin{minipage}{0.1\textwidth}
		\centering
		\footnotesize{Jigsaw-ViT}
		\end{minipage}\hfill
		\begin{minipage}{0.9\textwidth}
		\centering
		\includegraphics[width=0.1\columnwidth,height=0.1\columnwidth]{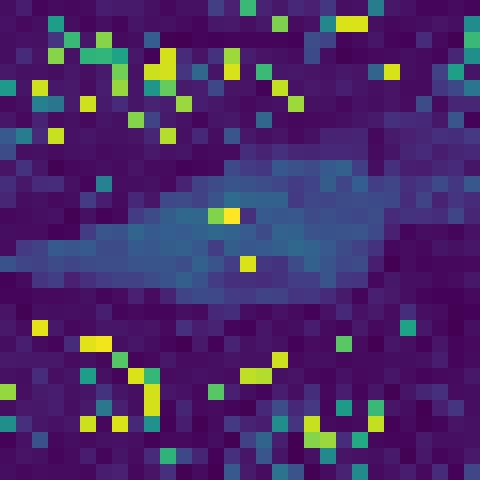}
		\includegraphics[width=0.1\columnwidth,height=0.1\columnwidth]{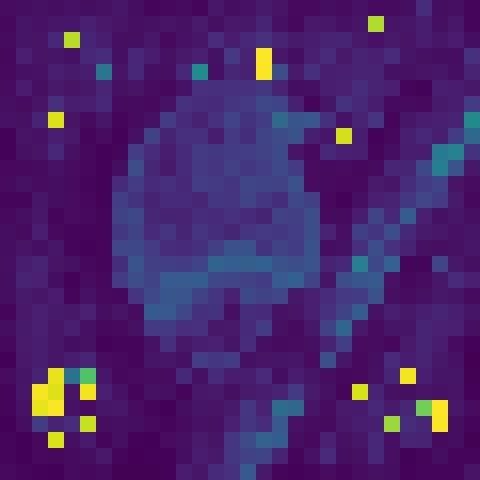}
		\includegraphics[width=0.1\columnwidth,height=0.1\columnwidth]{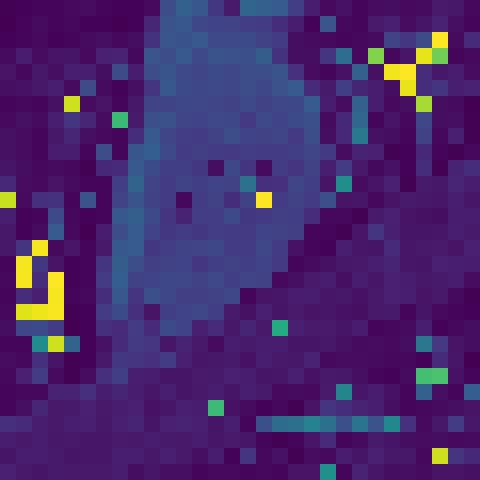}
		\includegraphics[width=0.1\columnwidth,height=0.1\columnwidth]{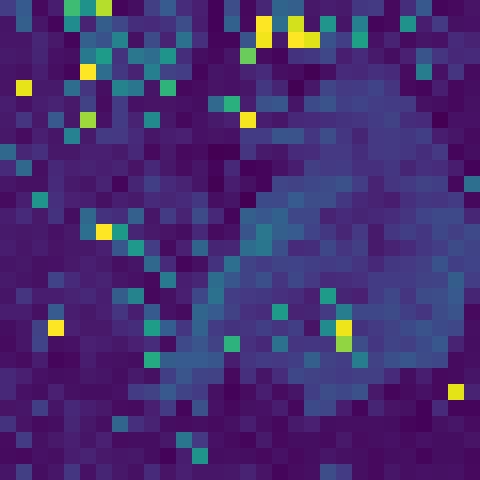}
		\includegraphics[width=0.1\columnwidth,height=0.1\columnwidth]{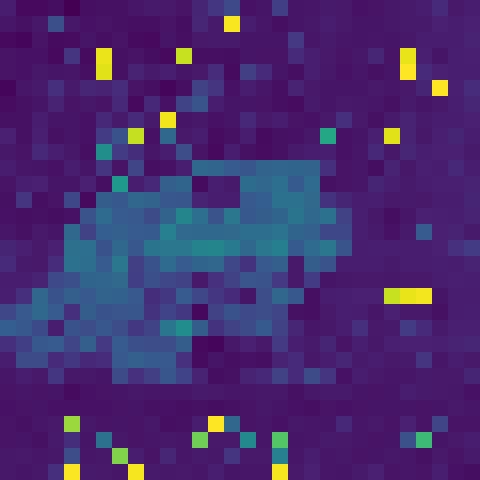}
		\includegraphics[width=0.1\columnwidth,height=0.1\columnwidth]{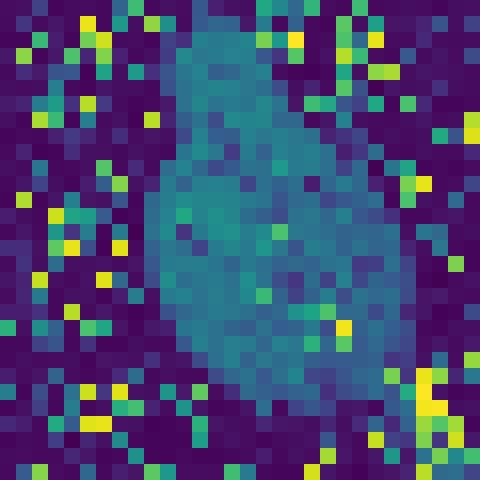}
		\includegraphics[width=0.1\columnwidth,height=0.1\columnwidth]{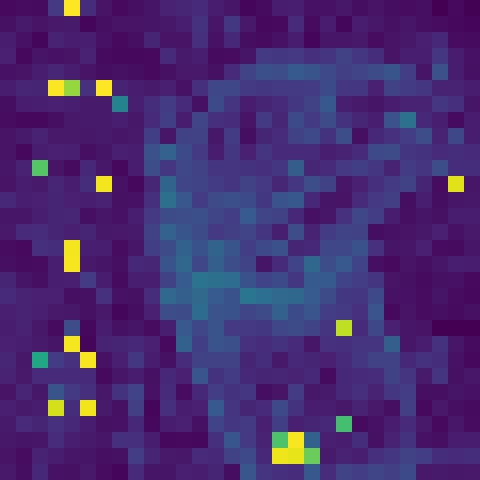}
		\includegraphics[width=0.1\columnwidth,height=0.1\columnwidth]{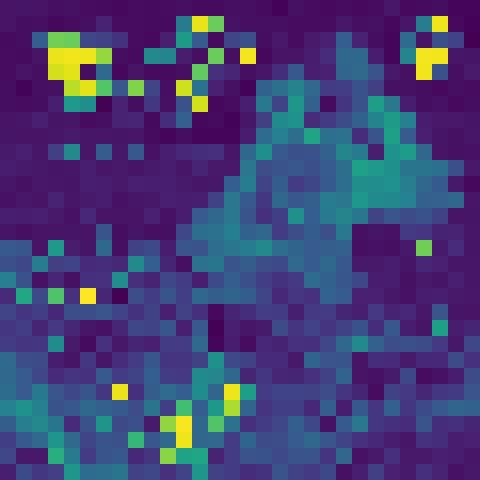}
		\includegraphics[width=0.1\columnwidth,height=0.1\columnwidth]{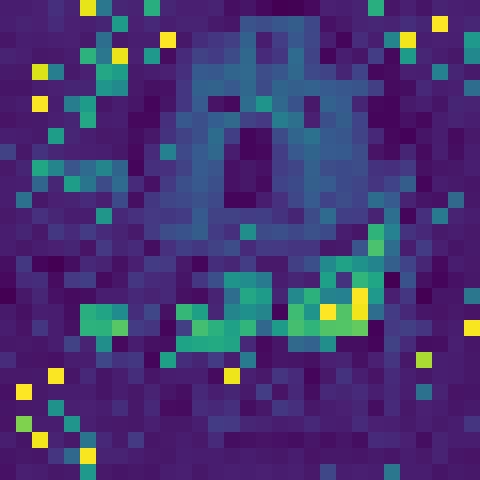}
		\end{minipage}
		\begin{minipage}{0.1\textwidth}
		\centering
		\footnotesize{Input}
		\end{minipage}\hfill
		\begin{minipage}{0.9\textwidth}
		\centering
		\includegraphics[width=0.1\columnwidth,height=0.1\columnwidth]{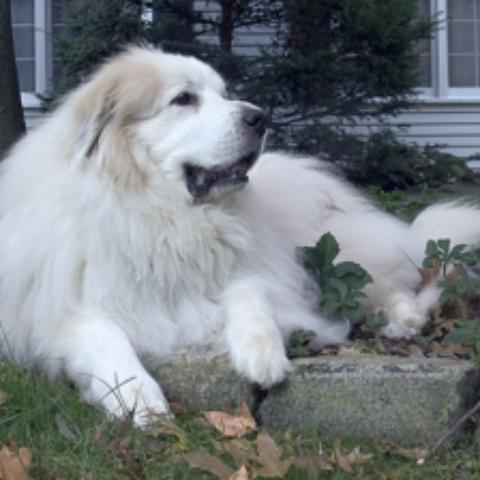}
		\includegraphics[width=0.1\columnwidth,height=0.1\columnwidth]{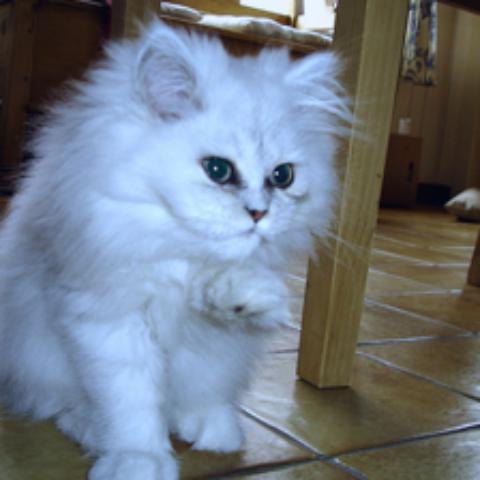}
		\includegraphics[width=0.1\columnwidth,height=0.1\columnwidth]{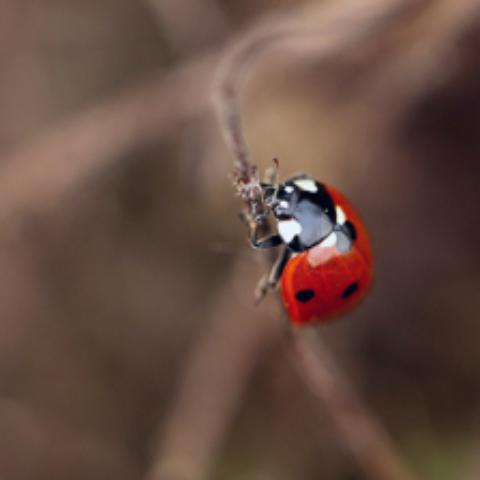}
		\includegraphics[width=0.1\columnwidth,height=0.1\columnwidth]{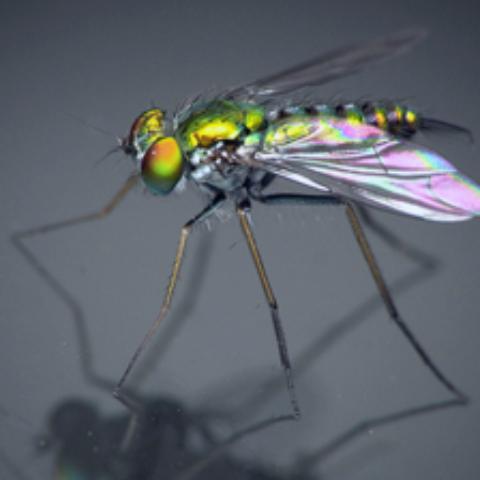}
		\includegraphics[width=0.1\columnwidth,height=0.1\columnwidth]{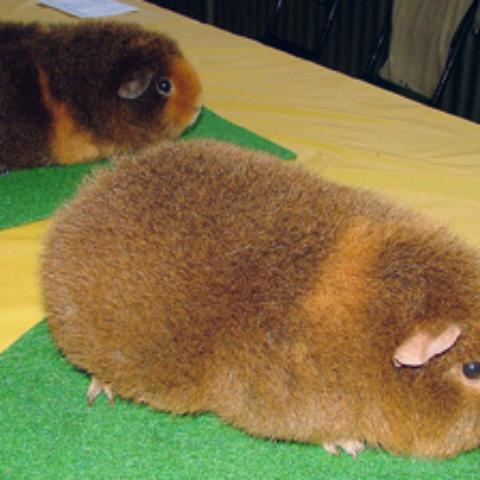}
		\includegraphics[width=0.1\columnwidth,height=0.1\columnwidth]{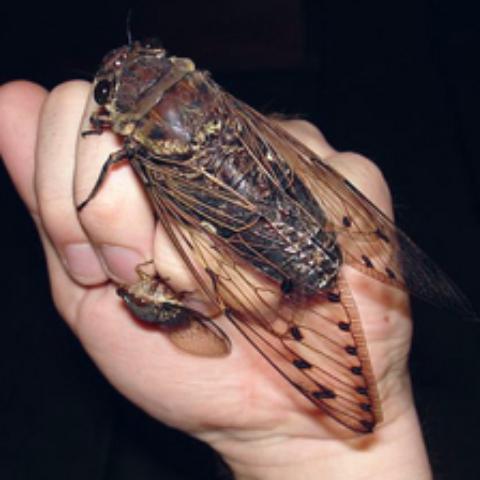}
		\includegraphics[width=0.1\columnwidth,height=0.1\columnwidth]{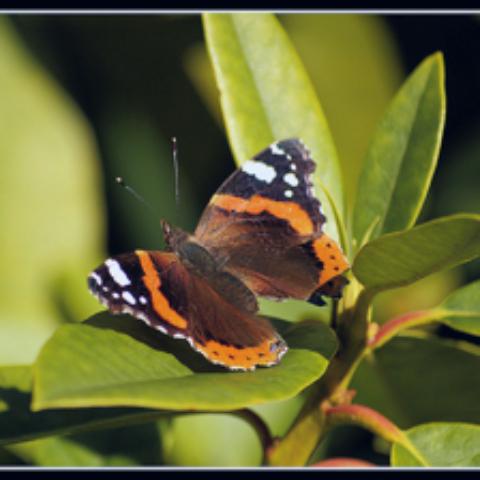}
		\includegraphics[width=0.1\columnwidth,height=0.1\columnwidth]{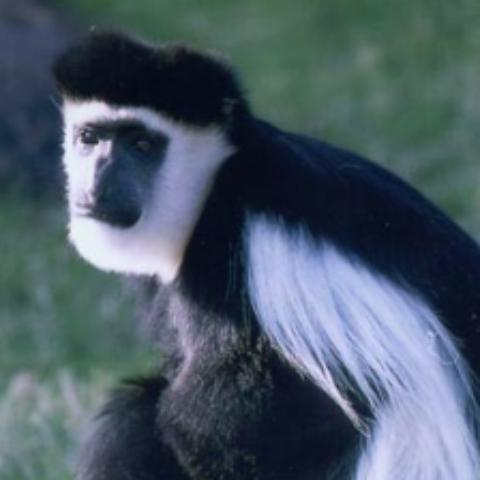}
		\includegraphics[width=0.1\columnwidth,height=0.1\columnwidth]{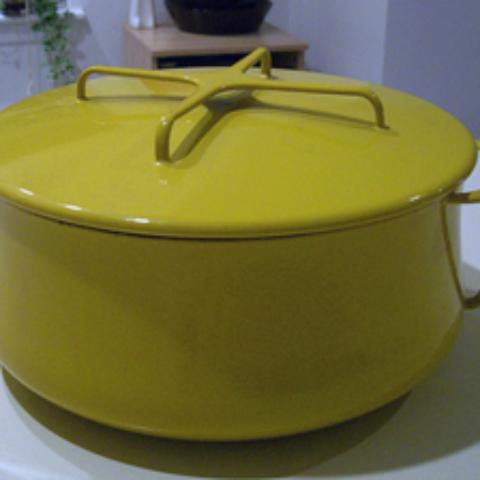}
		\end{minipage}\hfill
		\begin{minipage}{0.1\textwidth}
		\centering
		\footnotesize{DeiT-Small/16}
		\end{minipage}\hfill
		\begin{minipage}{0.9\textwidth}
		\centering
		\includegraphics[width=0.1\columnwidth,height=0.1\columnwidth]{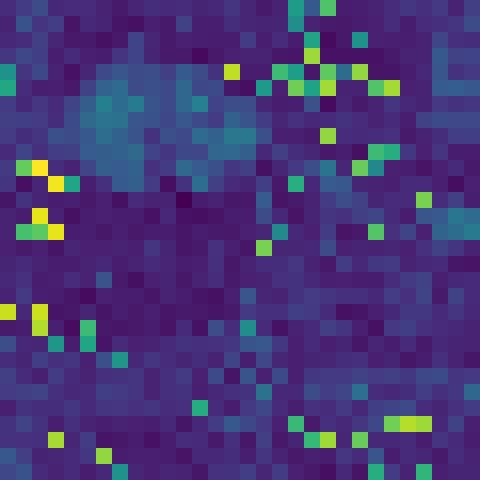}
		\includegraphics[width=0.1\columnwidth,height=0.1\columnwidth]{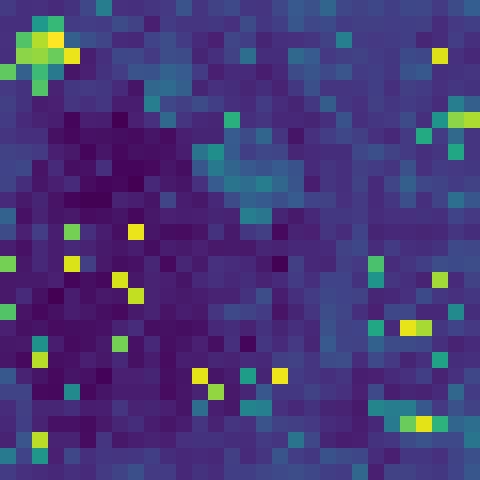}
		\includegraphics[width=0.1\columnwidth,height=0.1\columnwidth]{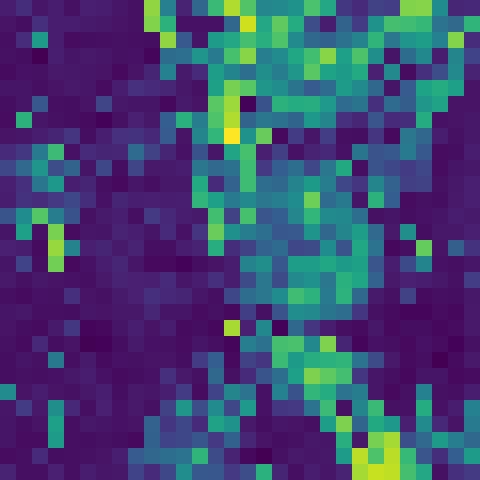}
		\includegraphics[width=0.1\columnwidth,height=0.1\columnwidth]{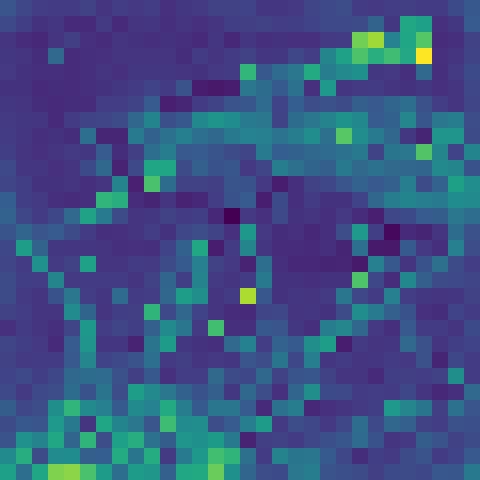}
		\includegraphics[width=0.1\columnwidth,height=0.1\columnwidth]{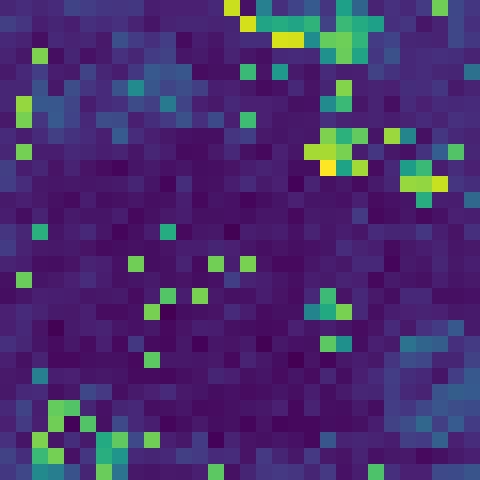}
		\includegraphics[width=0.1\columnwidth,height=0.1\columnwidth]{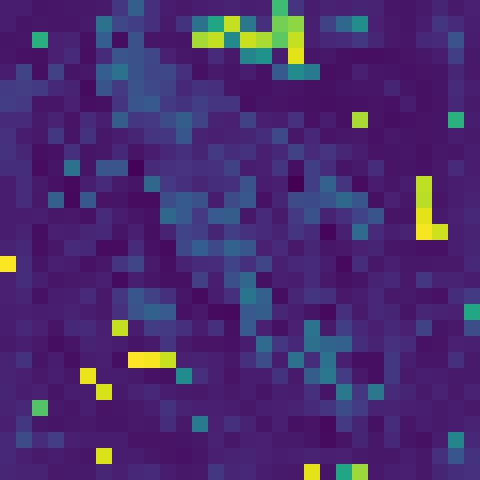}
		\includegraphics[width=0.1\columnwidth,height=0.1\columnwidth]{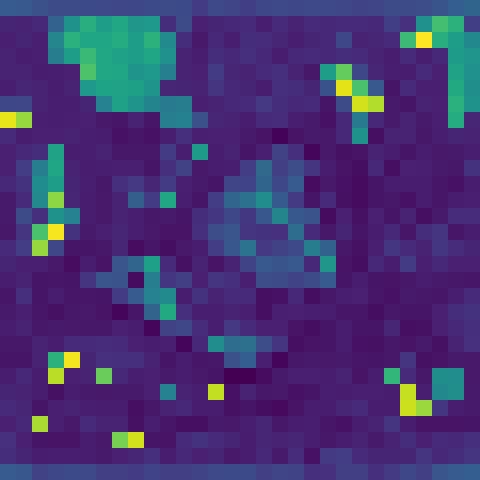}
		\includegraphics[width=0.1\columnwidth,height=0.1\columnwidth]{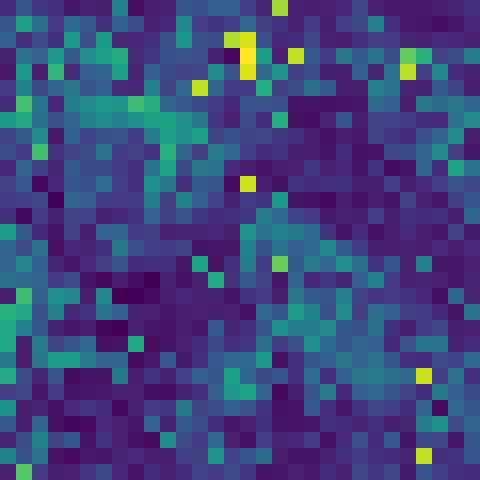}
		\includegraphics[width=0.1\columnwidth,height=0.1\columnwidth]{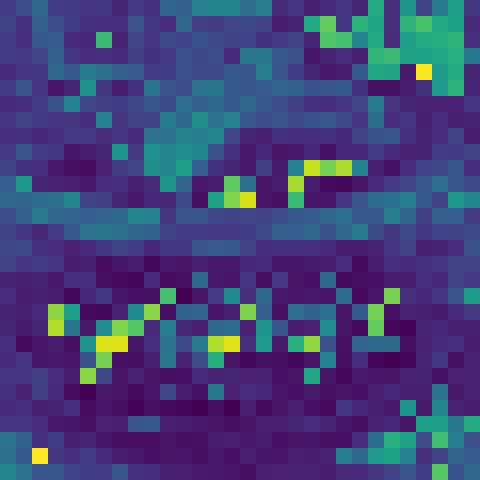}
		\end{minipage}\hfill
		\begin{minipage}{0.1\textwidth}
		\centering
		\footnotesize{Jigsaw-ViT}
		\end{minipage}\hfill
		\begin{minipage}{0.9\textwidth}
		\centering
		\includegraphics[width=0.1\columnwidth,height=0.1\columnwidth]{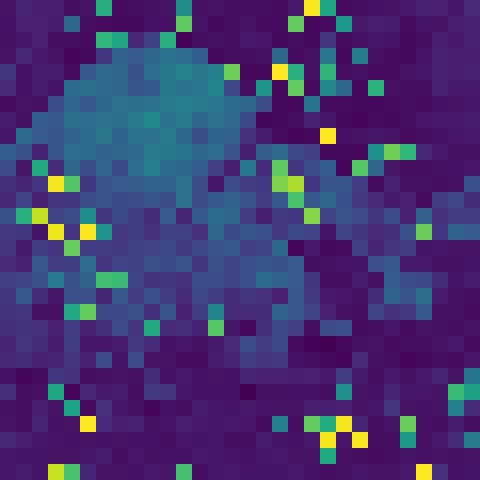}
		\includegraphics[width=0.1\columnwidth,height=0.1\columnwidth]{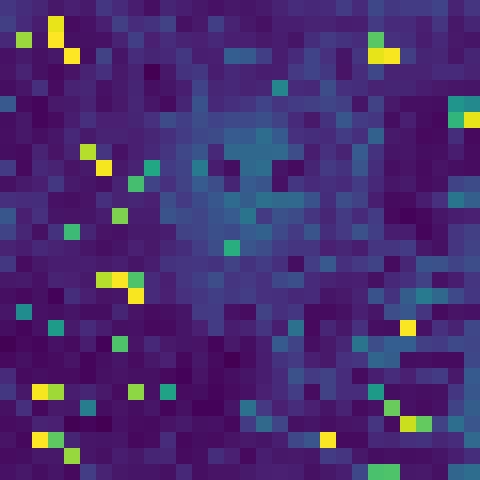}
		\includegraphics[width=0.1\columnwidth,height=0.1\columnwidth]{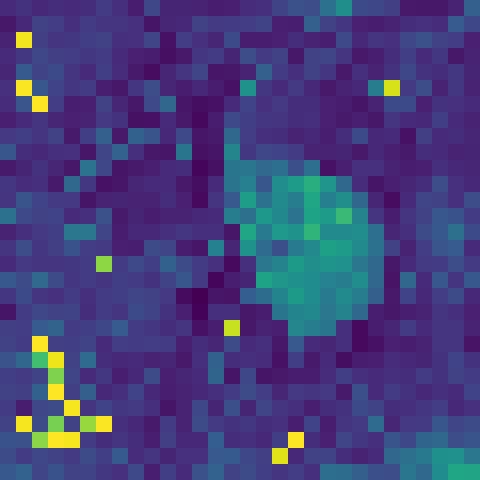}
		\includegraphics[width=0.1\columnwidth,height=0.1\columnwidth]{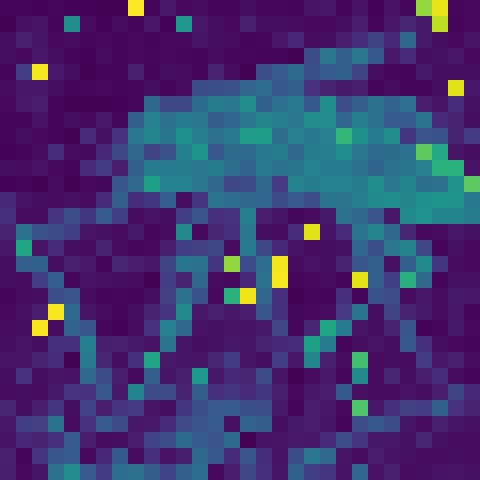}
		\includegraphics[width=0.1\columnwidth,height=0.1\columnwidth]{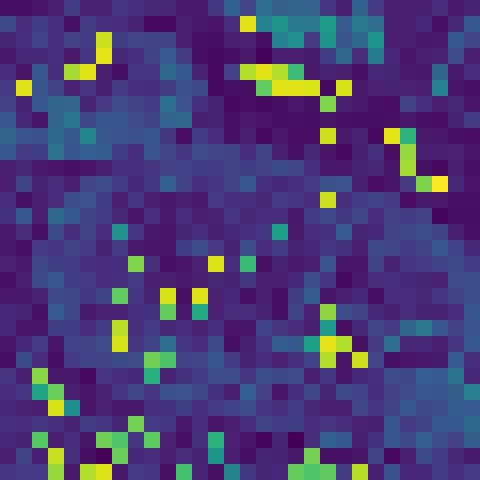}
		\includegraphics[width=0.1\columnwidth,height=0.1\columnwidth]{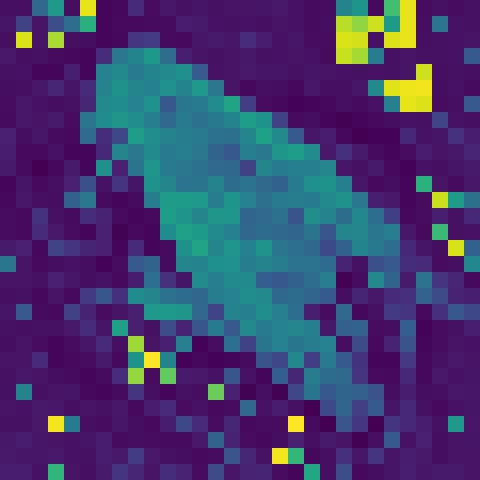}
		\includegraphics[width=0.1\columnwidth,height=0.1\columnwidth]{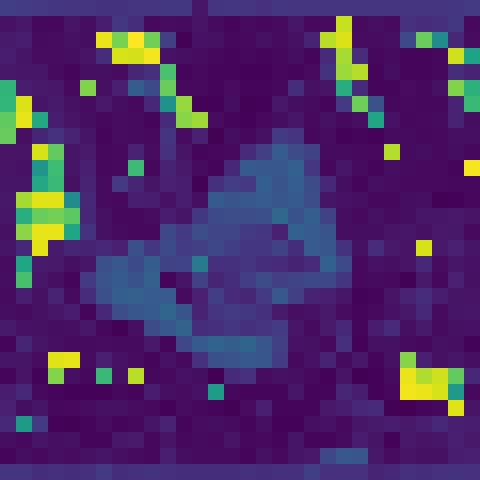}
		\includegraphics[width=0.1\columnwidth,height=0.1\columnwidth]{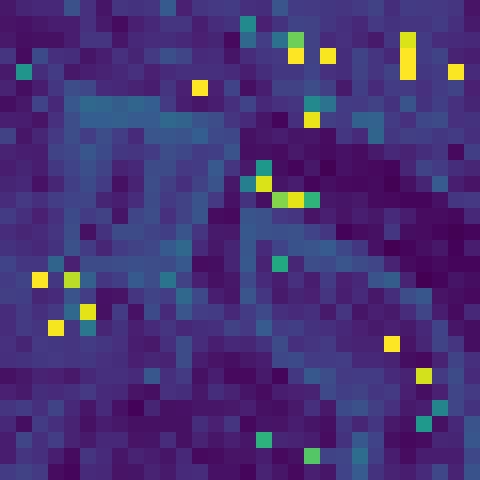}
		\includegraphics[width=0.1\columnwidth,height=0.1\columnwidth]{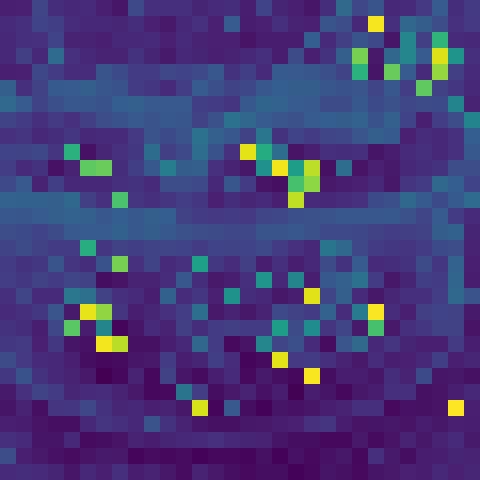}
	\end{minipage}
	\caption{{\bf Attention maps associated to the class token of the last layer.} We show the attention map for DeiT-Small/16 and Jigsaw-ViT trained on ImageNet-1K. Jigsaw-ViT learns more distinctive and informative salient-object attentions over the listed instances, which is in fact also the case for most of the training images.}
    \label{fig::supp-attention}
\end{figure*}

Similar as Fig.~\ref{fig::attention} in the main paper, here in Fig.~\ref{fig::supp-attention}, Jigsaw-ViT is able to learn more distinctive salient-object attentions than DeiT-Small/16.
Though we still observe noisy highlights on the background for all attention maps, Jigsaw-ViT provide more clearer fore-ground back-ground contrast than its counterpart DeiT-Small/16.

\section*{3\quad Additional implementation details}
\begin{figure}[t]
	\centering    
	{\includegraphics[width=\textwidth, height=\textwidth]{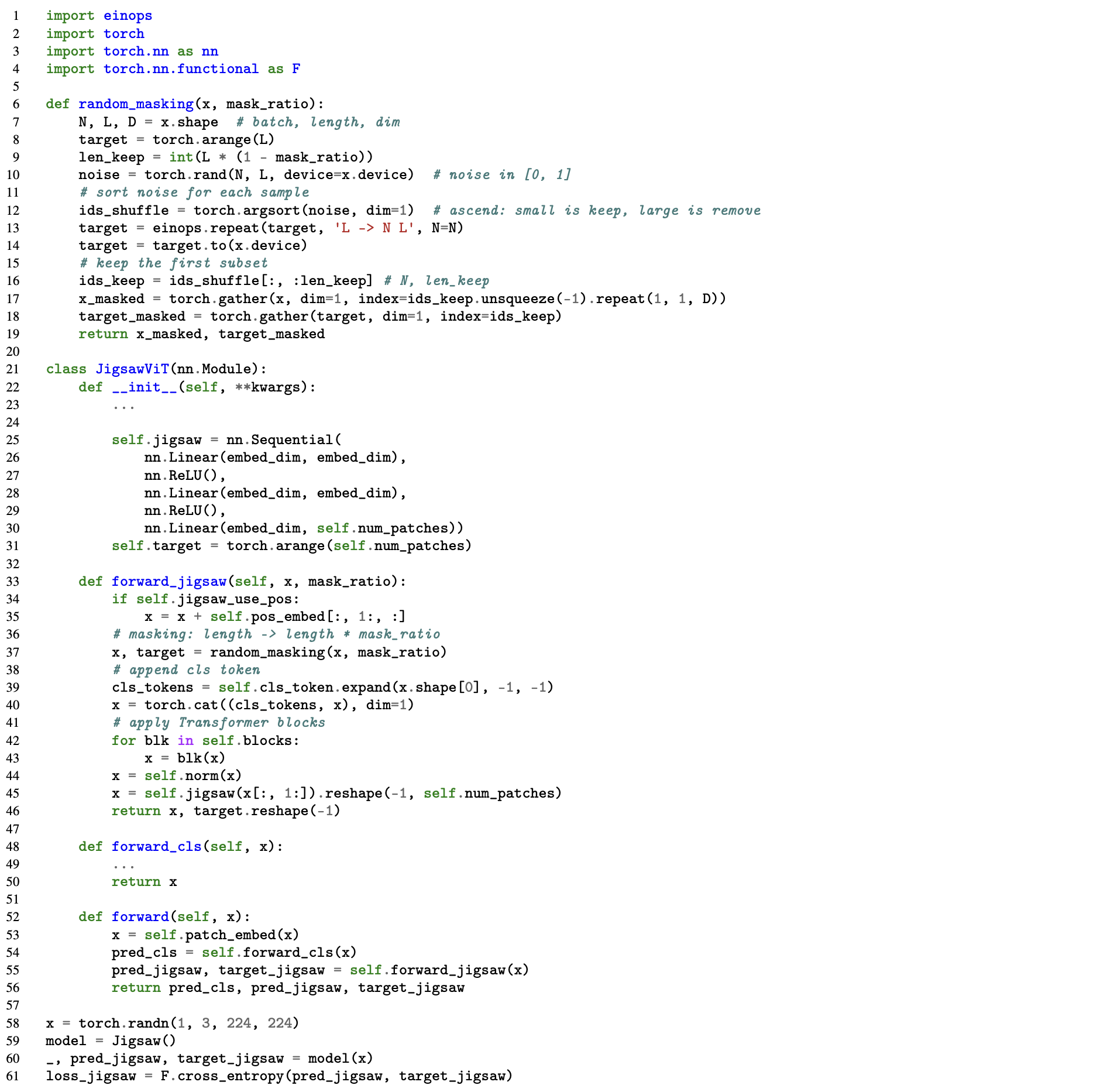}}\hfill
	\caption{\textbf{A PyTorch-like pseudo-code of our Jigsaw-ViT.} The entire code to reproduce the experiments will be made available online at our project page \url{https://yingyichen-cyy.github.io/Jigsaw-ViT}.}
	\label{fig::code}
\end{figure}

Our Jigsaw-ViT is mainly built on two public and official projects\footnote{\url{https://github.com/google-research/vision\_transformer}} and DeiT\footnote{\url{https://github.com/facebookresearch/deit}}.
To utilize the jigsaw flow during the training, a simple MLP module shall be sufficient to boost the accuracy of the standard ViT, as shown in Table~\ref{tab::mlp_head}.
In fact, although a higher accuracy can very likely be obtained by tuning both the number of MLP layers and the dimension of linear layers, our goal is to show that our proposed jigsaw flow is an easy-to-use approach that can can flexibly plugged into the optimization objective while without requiring tuning the module. 
In Fig.~\ref{fig::code}, we show the PyTorch-like pseudo-code for our proposed Jigsaw-VIT. In our experiments, the codes run with Python 3.6+, PyTorch 1.7.1 and Torchvision 0.8.2.

\captionsetup[table]{farskip=2pt,captionskip=1pt,aboveskip=4pt}
    \begin{table}[t]
    \renewcommand{\arraystretch}{1.1}
    \begin{center}
    \caption{\textbf{The details of the Jigsaw MLP head.} $d$ is the dimension of a token embedding, and  $o$ refers to the number of outputs, i.e., the number of patches in the input image.}
        \scalebox{0.8}{ 
            \begin{tabular}{lccc}
            \textbf{Layer} & \textbf{Activation} & \textbf{Output dimension}\\
            \hline
            Input & - & $d$\\
            \hline
            Linear & ReLU & $d$\\
            Linear & ReLU & $d$\\
            Linear & Softmax & $o$\\
            \hline
        \end{tabular}}
        \label{tab::mlp_head}
    \end{center}
    \end{table}
    
\end{document}